\documentclass[twoside,11pt]{article}
\usepackage{graphicx}
\usepackage{amsmath}
\usepackage{amssymb}
\usepackage{amsfonts}
\newcommand{\mb}{\mathbf}
\newcommand{\mc}{\mathcal}
\newcommand{\tb}{\textbf}
\newcommand{\ti}{\textit}
\newcommand{\bs}{\boldsymbol}
\usepackage{multirow}
\usepackage{pbox}
\usepackage{wrapfig}
\usepackage{algorithm}
\usepackage{algorithmic}
\makeatletter
\newcommand*{\rom}[1]{\expandafter\@slowromancap\romannumeral #1@}
\makeatother
\usepackage{color}

\usepackage{jmlr2e}
 
\jmlrheading{}{}{}{}{}{}{Pengtao Xie, Jun Zhu and Eric P. Xing}
 
\ShortHeadings{Diversity-Promoting Bayesian Learning of Latent Variable Models}{Xie, Zhu and Xing}
\firstpageno{1}
 
\begin{document}
 
\title{Diversity-Promoting Bayesian Learning of Latent Variable Models}
 
\author{\name Pengtao Xie \email pengtaox@cs.cmu.edu \\
       \addr Machine Learning Department\\
       Carnegie Mellon University\\
       Pittsburgh, PA 15213, USA\\
       \AND
        \name Jun Zhu \email dcszj@tsinghua.edu.cn \\
       \addr State Key Lab of Intelligent Technology and Systems\\
Tsinghua National Lab for Information Science and Technology\\
Department of Computer Science and Technology\\
Tsinghua University\\
Beijing, 100084 China\\
       \AND
        \name Eric P. Xing \email eric.xing@petuum.com \\
       \addr Petuum Inc.\\
       2555 Smallman St.\\
       Pittsburgh, PA 15222, USA\\
       \AND
}
 
\editor{}
 
\maketitle
%
\begin{abstract}
To address three important issues involved in latent variable models (LVMs), including capturing infrequent patterns, achieving small-sized but expressive models and alleviating overfitting, several studies have been devoted to ``diversifying'' LVMs, which aim at encouraging the components in LVMs to be diverse. Most existing studies fall into a frequentist-style regularization framework, where the components are learned via point estimation. In this paper, we investigate how to ``diversify'' LVMs in the paradigm of Bayesian learning. We propose two approaches that have complementary advantages. One is to define a diversity-promoting mutual angular prior which assigns larger density to components with larger mutual angles and use this prior to affect the posterior via Bayes' rule. We develop two efficient approximate posterior inference algorithms based on variational inference and MCMC sampling. The other approach is to impose diversity-promoting regularization directly over the post-data distribution of components. We also extend our approach to ``diversify'' Bayesian nonparametric models where the number of components is infinite. A sampling algorithm based on slice sampling and Hamiltonian Monte Carlo is developed. We apply these methods to ``diversify'' Bayesian mixture of experts model and infinite latent feature model. Experiments on various datasets demonstrate the effectiveness and efficiency of our methods.
\end{abstract}
 
\begin{keywords}
Diversity-Promoting, Latent Variable Models, Mutual Angular Prior, Bayesian Learning, Variational Inference
\end{keywords}
 
\section{Introduction}
Latent variable models (LVMs) \citep{bishop1998latent,knott1999latent,blei2014build} are a major workhorse in machine learning (ML) to extract latent \textit{patterns} underlying data, such as \textit{themes} behind documents and \textit{motifs} hiding in genome sequences.
To properly capture these patterns, LVMs are equipped with a set of \textit{components}, each of which is aimed to capture one pattern and is usually parametrized by a vector. For instance, in topic models \citep{blei2003latent}, each component (referred to as \textit{topic}) is in charge of capturing one \textit{theme} underlying documents and is represented by a multinomial vector.

While existing LVMs have demonstrated great success, they are less capable in addressing two new problems emerged due to the growing volume and complexity of data.
First, it is often the case that the frequency of patterns is distributed in a power-law fashion \citep{wang2014peacock,xie2015diversifying} where a handful of patterns occur very frequently whereas most patterns are of low frequency. Existing LVMs lack capability to capture infrequent patterns, which is possibly due to the design of LVMs' objective function used for training. For example, a maximum likelihood estimator would reward itself by modeling the frequent patterns well as they are the major contributors of the likelihood function. On the other hand, infrequent patterns contribute much less to the likelihood, thereby it is not very rewarding to model them well and LVMs tend to ignore them.
Infrequent patterns often carry valuable information, thus should not be ignored. For instance, in a topic modeling based recommendation system, an infrequent topic (pattern) like \textit{losing weight} is more likely to improve the click-through rate than a frequent topic like \textit{politics}. 
Second, the number of components $K$ strikes a tradeoff between model size (complexity) and modeling power. For a small $K$, the model is not expressive enough to sufficiently capture the complex patterns behind data; for a large $K$, the model would be of large size and complexity, incurring high computational overhead. How to reduce model size while preserving modeling power is a challenging issue.

To cope with the two problems, several studies \citep{Zou_priorsfor,xie2015diversifying,xie2015learning} propose a ``diversification'' approach, which encourages the components of a LVM to be mutually ``dissimilar''. First, regarding capturing infrequent patterns, as posited in \citep{xie2015diversifying} ``diversified'' components are expected to be less aggregated over frequent patterns and part of them would be spared to cover the infrequent patterns. 
Second, concerning shrinking model size without compromising modeling power, \citet{xie2015learning} argued that ``diversified'' components bear less redundancy and are mutually complementary, making it possible to capture information sufficiently well with a small set of components, i.e., obtaining LVMs possessing high representational power and low computational complexity.

The existing studies \citep{Zou_priorsfor,xie2015diversifying,xie2015learning} of ``diversifying'' LVMs mostly focus on {\it point estimation} \citep{wasserman2013all} of the model components, under a frequentist-style regularized optimization framework. In this paper, we study how to promote diversity under an alternative learning paradigm: Bayesian inference \citep{jaakkola1997variational,bishop2003bayesian,
neal2012bayesian}, where the components are considered as random variables of which a {\it posterior distribution} shall be computed from data under certain priors. Compared with point estimation, Bayesian learning offers complementary benefits. First, it offers a ``model-averaging" \citep{jaakkola1997variational,bishop2003bayesian} effect for LVMs when they are used for decision-making and prediction because the parameters shall be integrated under a posterior distribution, thus potentially alleviate overfitting on training data. Second, it provides a natural way to quantify uncertainties of model parameters, and downstream decisions and predictions made thereupon \citep{jaakkola1997variational,bishop2003bayesian,neal2012bayesian}. \citet{affandi2013approximate} investigated the ``diversification'' of Bayesian LVMs using the determinantal point process (DPP) \citep{kulesza2012determinantal} prior. While Markov chain Monte Carlo (MCMC)~\citep{affandi2013approximate} methods have been developed for approximate posterior inference under the DPP prior, DPP is not amenable for another mainstream paradigm of approximate inference techniques -- variational inference \citep{wainwright2008graphical} -- which is usually more efficient \citep{hoffman2013stochastic} than MCMC. In this paper, we propose alternative diversity-promoting priors that overcome this limitation.

We propose two approaches that have complementary advantages to perform diversity-promoting Bayesian learning of LVMs. Following \citep{xie2015diversifying}, we adopt a notion of diversity that component vectors are more diverse provided the pairwise angles between them are larger.
First, we define mutual angular Bayesian network (MABN) priors over the components, which assign higher probability density to components that have larger mutual angles and use these priors to affect the posterior via Bayes' rule.
Specifically, we build a Bayesian network \citep{koller2009probabilistic} whose nodes represent the directional vectors of the components and local probabilities are parameterized by von Mises-Fisher \citep{mardia2009directional} distributions that entail an inductive bias towards vectors with larger mutual angles. The MABN priors are amenable for approximate posterior inference of model components. In particular, they facilitate variational inference, which is usually more efficient than MCMC sampling.
Second, in light of that it is not flexible (or even possible) to define priors to capture certain diversity-promoting effects such as small variance of mutual angles, we adopt a posterior regularization approach \citep{zhu2014bayesian}, in which a diversity-promoting regularizer is directly imposed over the post-data distributions to encourage diversity and the regularizer can be flexibly defined to accommodate various desired diversity-promoting goals. We instantiate the two approaches to the Bayesian mixture of experts model (BMEM) \citep{waterhouse1996bayesian} and experiments demonstrate the effectiveness and efficiency of our approaches.
 
We also study how to ``diversify'' Bayesian nonparametric LVMs (BN-LVMs)~\citep{ferguson1973bayesian,ghahramani2005infinite,hjort2010bayesian}. Different from parametric LVMs where the component number is set to an finite value and does not change throughout the entire execution of algorithm, in BN-LVMs the number of components is unlimited and can reach infinite in principle. As more data accumulates, new components are dynamically added. Compared with parametric LVMs, BN-LVMs possess the following advantages: (1) they are highly flexible and adaptive: if new data cannot be well modeled by existing components, new components are automatically invoked; (2) in BN-LVMs, the ``best" number of components is determined according to the fitness to data, rather than being manually set which is a challenging task even for domain experts. To ``Diversify" BN-LVMs, we extend the MABN prior to an Infinite Mutual Angular (IMA) prior that encourages infinitely many components to have large angles. In this prior, the components are mutually dependent, which incurs great challenges for posterior inference. We develop an efficient sampling algorithm based on slice sampling~\citep{tehstick} and Riemann manifold Hamiltonian Monte Carlo~\citep{girolami2011riemann}. We apply the IMA prior to induce diversity in the infinite latent feature model (ILFM) \citep{ghahramani2005infinite} and experiments on various datasets demonstrate that the IMA is able to (1) achieve better performance with fewer components; (2) better capture infrequent patterns; and (3) reduce overfitting.
 
The major contributions of this work are:
\begin{itemize}
\setlength\itemsep{0.1em}
\item We propose a mutual angular Bayesian network (MABN) prior which is biased towards components having large mutual angles, to promote diversity in Bayesian LVMs.
\item We develop an efficient variational inference method for posterior inference of model components under the MABN priors.
\item To flexibly accommodate various diversity-promoting effects, we study a posterior regularization approach which directly imposes diversity-promoting regularization over the post-data distributions.
\item We extend the MABN prior from the finite case to the infinite case and apply it to ``diversify" Bayesian nonparametric models. 
\item We develop an efficient sampling algorithm based on slice sampling and Riemann manifold Hamiltonian Monte Carlo for ``diversified'' BN-LVMs.
\item Using Bayesian mixture of experts model and infinite latent feature model as study cases, we empirically demonstrate the effectiveness and efficiency of our methods.
\end{itemize}
The rest of the paper is organized as follows. Section 2 reviews related works. In Section 3 and 4, we introduce how to promote diversity in Bayesian parametric and nonparametric LVMs respectively. Section 5 gives experimental results and Section 6 concludes the paper.

\section{Related Work}

Recent works \citep{Zou_priorsfor,xie2015diversifying,xie2015learning} have studied the diversification of components in LVMs under a point estimation framework. \citet{Zou_priorsfor} leverage the determinantal point process (DPP) \citep{kulesza2012determinantal} to promote diversity in latent variable models.
\citet{xie2015diversifying} propose a mutual angular regularizer that encourages model components to be mutually different where the dissimilarity is measured by angles. \citet{cogswell2015reducing} define a covariance-based regularizer to reduce the correlation among hidden units in neural networks, for the sake of alleviating overfitting.
 
Diversity-promoting Bayesian learning of LVMs has been investigated in \citep{affandi2013approximate}, which utilizes the DPP prior to induce bias towards diverse components. They develop a Gibbs sampling \citep{gilks2005markov} algorithm. But the determinant in DPP makes variational inference based algorithms very difficult to derive. Our conference version of the paper~\citep{xie2016diversity} has introduced a mutual angular prior to ``diversify'' Bayesian parametric LVMs. This work extends the study of ``diversification'' to nonparametric models where the number of components is infinite.
 
Diversity-promoting regularization is investigated in other problems as well, such as ensemble learning and classification. In ensemble learning, many studies \citep{kuncheva2003measures,banfield2005ensemble,partalas2008focused,yu2011diversity} explore how to select a diverse subset of base classifiers or regressors, with the aim to improve generalization performance and reduce computational complexity. In multi-way classification, \citet{malkin2008ratio} propose to use the determinant of a covariance matrix to encourage ``diversity'' among classifiers. \citet{jalali2015variational} propose a class of \ti{variational Gram functions} (VGFs) to promote pairwise dissimilarity among classifiers.
 
\section{Diversity-Promoting Bayesian Learning of Parametric Latent Variable Models}

In this section, we study how to ``diversify'' parametric Bayesian LVMs where the number of components is finite. We investigate two approaches: prior control and posterior regularization, which have complementary advantages.
 
 
\subsection{Diversity-Promoting Mutual Angular Prior}
\label{sec:map}
The first approach we take is to define a prior which has an inductive bias towards components that are more ``diverse'' and use it to affect the posterior via Bayes' rule. We refer to this approach as \textit{prior control}. While diversity can be defined in various ways, following \citep{xie2015diversifying} we adopt the notion that a set of component vectors are considered to be more diverse if the pairwise angles between them are larger.
We desire the prior to have two traits. First, to favor diversity, they assign a higher density to components having larger mutual angles. Second, it should facilitate posterior inference. In Bayesian learning, the easiness of posterior inference relies heavily on the prior \citep{blei2006correlated,wang2013variational}.
 
One possible solution is to turn the mutual angular regularizer $\Omega(\mb{A})$ \citep{xie2015diversifying} that encourages a set of component vectors $\mb{A}=\{\mb{a}_i\}_{i=1}^{K}$ to have large mutual angles into a distribution $p(\mb{A})=\frac{1}{Z}\exp(\Omega(\mb{A}))$ based on Gibbs measure \citep{kindermann1980markov}, where $Z$ is the partition function guaranteeing that $p(\mb{A})$ integrates to one. The concern is that it is not sure whether $Z=\int_{\mb{A}} \exp(\Omega(\mb{A})) \mathrm{d}\mb{A}$ is finite, i.e., whether $p(\mb{A})$ is proper. When an improper prior is utilized in Bayesian learning, the posterior is also highly likely to be improper, except in a few special cases \citep{wasserman2013all}. Performing inference on improper posteriors is problematic.
 
\begin{figure}
\begin{center}
\includegraphics[width=0.45\columnwidth]{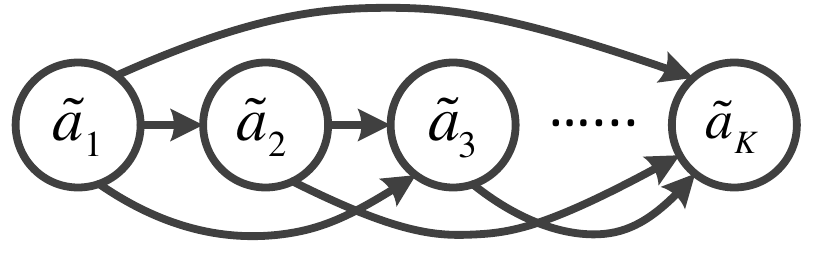}
\end{center}
 
\caption{A Bayesian Network Representation of the Mutual Angular Prior}
 
\label{fig:bn}
\end{figure}

Here we define mutual angular Bayesian network (MABN) priors possessing the aforementioned two traits, based on Bayesian network \citep{koller2009probabilistic} and von Mises-Fisher \citep{mardia2009directional} distribution. For technical convenience, we decompose each real-valued component vector $\mb{a}$ into $\mb{a}=g\mb{\tilde{a}}$, where $g=\|\mb{a}\|_2$ is the magnitude and $\mb{\tilde{a}}$ is the direction ($\|\mb{\tilde{a}}\|_2=1$). Let $\mb{\widetilde{A}}=\{\mb{\tilde{a}}_{i}\}_{i=1}^{K}$ denote the directional vectors. Note that the angle between two vectors is invariant to their magnitudes, thereby, the mutual angles of component vectors in $\mb{A}$ are the same as angles of directional vectors in $\mb{\widetilde{A}}$.
We first construct a prior which prefers vectors in $\mb{\widetilde{A}}$ to possess large angles. The basic idea is to use a Bayesian network (BN) to characterize the dependency among directional vectors and design local probabilities to entail an inductive bias towards large mutual angles. In the Bayesian network (BN) shown in Figure \ref{fig:bn}, each node $i$ represents a directional vector $\mb{\tilde{a}}_i$ and its parents $\textrm{pa}(\mb{\tilde{a}}_i)$ are nodes $1,\cdots, i-1$. We define a local probability at node $i$ to encourage $\mb{\tilde{a}}_i$ to have large mutual angles with $\mb{\tilde{a}}_1,\cdots,\mb{\tilde{a}}_{i-1}$. Since these directional vectors lie on a sphere, we use the von Mises-Fisher (vMF) distribution to model them. The probability density function of the vMF distribution is $f(\mb{x})=C_{p}(\kappa)\exp(\kappa\bs\mu^\top \mb{x})$, where the random variable $\mb{x}\in\mathbb{R}^p$ lies on a $p-1$ dimensional sphere ($\|\mb{x}\|_2=1$), $\bs\mu$ is the mean direction with $\|\bs\mu\|_2=1$, $\kappa>0$ is the concentration parameter and $C_{p}(\kappa)$ is the normalization constant.
The local probability $p(\mb{\tilde{a}}_i|\textrm{pa}(\mb{\tilde{a}}_i))$ at node $i$ is defined as a von Mises-Fisher (vMF) distribution whose density is
\setlength\arraycolsep{-2pt} \begin{eqnarray} \label{eq:map1}
&& p(\mb{\tilde{a}}_i|\textrm{pa}(\mb{\tilde{a}}_i))=C_{p}(\kappa)\exp\left( \kappa\left(-\frac{\sum_{j=1}^{i-1}\tilde{\mb{a}}_j}{\|\sum_{j=1}^{i-1}\tilde{\mb{a}}_j\|_2}\right)^\top \tilde{\mb{a}}_i \right)
\end{eqnarray}
with mean direction $-\sum_{j=1}^{i-1}\mb{\tilde{a}}_j/||\sum_{j=1}^{i-1}\mb{\tilde{a}}_j||_2$. 
 
Now we explain why this local probability favors large mutual angles. Since $\mb{\tilde{a}}_i$ and $\mb{\tilde{a}}_j$ are unit-length vectors, $\mb{\tilde{a}}_j^\top \mb{\tilde{a}}_i$ is the cosine of the angle between $\mb{\tilde{a}}_i$ and $\mb{\tilde{a}}_j$. If $\mb{\tilde{a}}_i$ has larger angles with $\{\mb{\tilde{a}}_j\}_{j=1}^{i-1}$, then the average negative cosine similarity $(-\sum_{j=1}^{i-1}\mb{\tilde{a}}_j)^\top\mb{\tilde{a}}_i$ would be larger, accordingly $p(\mb{\tilde{a}}_i|\textrm{pa}(\mb{\tilde{a}}_i))$ would be larger. This statement is true for all $i>1$. As a result, $p(\mb{\widetilde{A}})=p(\mb{\tilde{a}}_1)\prod_{i=2}^{K}p(\mb{\tilde{a}}_i|\textrm{pa}(\mb{\tilde{a}}_i))$ would be larger if the directional vectors have larger mutual angles. For the magnitudes $\{g_i\}_{i=1}^K$ of the components, which have nothing to do with the mutual angles, we sample $g_i$ for each component independently from a gamma distribution with shape parameter $\alpha_1$ and rate parameter $\alpha_2$.
The generative process of $\mb{A}$ is summarized as follows:
\begin{itemize}
\setlength\itemsep{0em}
\item Draw $\mb{\tilde{a}}_1\sim \textrm{vMF}(\bs\mu_{0},\kappa)$
\item For $i=2,\cdots,K$,
draw $\mb{\tilde{a}}_i\sim \textrm{vMF}(-\frac{\sum_{j=1}^{i-1}\tilde{\mb{a}}_j}{\|\sum_{j=1}^{i-1}\tilde{\mb{a}}_j\|_2},\kappa)$
\item For $i=1,\cdots,K$, draw $g_i\sim \textrm{Gamma}(\alpha_1,\alpha_2)$
\item For $i=1,\cdots,K$, let $\mb{a}_i=\mb{\tilde{a}}_i g_i$
\end{itemize}
The probability distribution over $\mb{A}$ can be written as
\setlength\arraycolsep{1pt}\begin{equation}
\label{eq:map_1}
\begin{array}{l}
p(\mb{A})=C_{p}(\kappa)\exp(\kappa \mu_{0}^\top \mb{\tilde{a}}_1)\prod_{i=2}^{K}C_{p}(\kappa)\exp\!\left(\kappa
\left(-\frac{\sum_{j=1}^{i-1}\tilde{\mb{a}}_j}{\|\sum_{j=1}^{i-1}\tilde{\mb{a}}_j\|_2} \right)^\top \!\! \tilde{\mb{a}}_i \right) \!\! \prod_{i=1}^{K} \! \frac{\alpha_2^{\alpha_1}g_i^{\alpha_1-1} e^{-g_i\alpha_2}}{\Gamma(\alpha_1)} .
\end{array}
\end{equation}
According to the factorization theorem \citep{koller2009probabilistic} of Bayesian network, it is easy to verify $\int_{\mb{A}} p(\mb{A}) \mathrm{d}\mb{A}=1$, thus $p(\mb{A})$ is a proper prior.
 
When inferring the posterior of model components using a variational inference method, we need to compute the expectation of $1/\|\sum_{j=1}^{i-1}\mb{\tilde{a}}_j||_2$ appearing in the local probability $p(\mb{\tilde{a}}_i|\textrm{pa}(\mb{\tilde{a}}_i))$, which is extremely difficult. To address this issue, we define an alternative local probability that achieves similar modeling effect as $p(\mb{\tilde{a}}_i|\textrm{pa}(\mb{\tilde{a}}_i))$, but greatly facilitates variational inference. We re-parametrize the local probability $\hat{p}(\mb{\tilde{a}}_i|\textrm{pa}(\mb{\tilde{a}}_i))$ defined in Eq.(\ref{eq:map1}) using Gibbs measure:
\begin{eqnarray}
\label{eq:map_2}
\hat{p}(\mb{\tilde{a}}_i|\textrm{pa}(\mb{\tilde{a}}_i))
&\propto& \exp\left( \kappa(-\sum_{j=1}^{i-1}\tilde{\mb{a}}_j)^\top \tilde{\mb{a}}_i \right) \nonumber \\
&\propto&\exp\left(\kappa\|\sum_{j=1}^{i-1}\tilde{\mb{a}}_j\|_2(-\frac{\sum_{j=1}^{i-1}\tilde{\mb{a}}_j}{\|\sum_{j=1}^{i-1}\tilde{\mb{a}}_j\|_2})^\top \tilde{\mb{a}}_i\right) \nonumber \\
&=&C_{p}\left(\kappa\|\sum_{j=1}^{i-1}\tilde{\mb{a}}_j\|_2\right)\exp\left(\kappa\|\sum_{j=1}^{i-1}\tilde{\mb{a}}_j\|_2(-\frac{\sum_{j=1}^{i-1}\tilde{\mb{a}}_j}{\|\sum_{j=1}^{i-1}\tilde{\mb{a}}_j\|_2})^\top \tilde{\mb{a}}_i\right) \nonumber \\
&=&C_{p}\left(\kappa\|\sum_{j=1}^{i-1}\tilde{\mb{a}}_j\|_2\right)\exp\left(\kappa(-\sum_{j=1}^{i-1}\tilde{\mb{a}}_j)^\top \tilde{\mb{a}}_i \right),
\end{eqnarray}
which is another vMF distribution with mean direction $-\sum_{j=1}^{i-1}\tilde{\mb{a}}_j/\|\sum_{j=1}^{i-1}\tilde{\mb{a}}_j\|_2$ and concentration parameter $\kappa\|\sum_{j=1}^{i-1}\tilde{\mb{a}}_j\|_2$. This re-parameterized local probability is proportional to $(-\sum_{j=1}^{i-1}\tilde{\mb{a}}_j)^\top \tilde{\mb{a}}_i$, which measures the negative cosine similarity between $\tilde{\mb{a}}_i$ and its parent vectors. Thereby, $\hat{p}(\mb{\tilde{a}}_i|\textrm{pa}(\mb{\tilde{a}}_i))$ still encourages large mutual angles between vectors as $p(\mb{\tilde{a}}_i|\textrm{pa}(\mb{\tilde{a}}_i))$ does. The difference between $\hat{p}(\mb{\tilde{a}}_i|\textrm{pa}(\mb{\tilde{a}}_i))$ and $p(\mb{\tilde{a}}_i|\textrm{pa}(\mb{\tilde{a}}_i))$ is that in $\hat{p}(\mb{\tilde{a}}_i|\textrm{pa}(\mb{\tilde{a}}_i))$ the term $\|\sum_{j=1}^{i-1}\tilde{a}_j\|_2$ is moved from the denominator to the normalizer, thus we can avoid computing the expectation of $1/\|\sum_{j=1}^{i-1}\tilde{a}_j\|_2$. Though it incurs a new problem that we need to compute the expectation of $\log C_{p}(\kappa\|\sum_{j=1}^{i-1}\tilde{\mb{a}}_j\|_2)$, which is also hard due to the complex form of the $C_{p}(\cdot)$ function, we managed to resolve this problem as detailed in Section \ref{sec:vi}. We refer to the MABN prior defined in Eq.(\ref{eq:map_1}) as type I MABN and that with local probability defined in Eq.(\ref{eq:map_2}) as type II MABN.

\subsubsection{Approximate Inference Algorithms}
\label{sec:vi}
We develop algorithms to infer the posteriors of components under the MABN prior. Since exact posteriors are intractable, we resort to approximate inference techniques. Two main paradigms of approximate inference methods are: (1) variational inference (VI) \citep{wainwright2008graphical}; (2) Markov chain Monte Carlo (MCMC) sampling \citep{gilks2005markov}. These two approaches possess benefits that are mutually complementary. MCMC can achieve a better approximation of the posterior than VI since it generates samples from the exact posterior while VI seeks an approximation. However, VI can be computationally more efficient \citep{hoffman2013stochastic}.

\paragraph{Variational Inference}

The basic idea of VI \citep{wainwright2008graphical} is to use a ``simpler'' variational distribution $q(\mb{A})$ to approximate the true posterior by minimizing the Kullback-Leibler divergence between these two distributions, which is equivalent to maximizing the following variational lower bound w.r.t $q(\mb{A})$:
\begin{equation}
\label{eq:vlb}
\begin{array}{l}
\mathbb{E}_{q(\mb{A})}[\log p(\mathcal{D}|\mb{A})]+\mathbb{E}_{q(\mb{A})}[\log p(\mb{A})]-\mathbb{E}_{q(\mb{A})}[\log q(\mb{A})]
\end{array}
\end{equation}
where $p(\mb{A})$ is the MABN prior and $p(\mathcal{D}|\mb{A})$ is data likelihood.
Here we choose $q(\mb{A})$ to be a mean field variational distribution $q(\mb{A})=
\prod_{k=1}^{K}q(\tilde{\mb{a}}_k)q(g_k)$, where $q(\tilde{\mb{a}}_k)=\textrm{vMF}(\tilde{\mb{a}}_k|\hat{\mb{a}}_k,\hat{\kappa})$ and $q(g_k)=\textrm{Gamma}(g_k|r_k,s_k)$. Given the variational distribution, we first compute the analytical expression of the variational lower bound, in which we particularly discuss how to compute $\mathbb{E}_{q(\mb{A})}[\log p(\mb{A})]$.
If choosing $p(\mb{A})$ to be the type-I MABN prior (Eq.(\ref{eq:map_1})), we need to compute $\mathbb{E}_{}[(-\frac{\sum_{j=1}^{i-1}\tilde{\mb{a}}_j}{\|\sum_{j=1}^{i-1}\tilde{\mb{a}}_j\|_2})^\top \tilde{\mb{a}}_i]$ which is very difficult to deal with due to the presence of $1/\|\sum_{j=1}^{i-1}\tilde{\mb{a}}_j\|_2$.
Instead we choose the type-II MABN prior for the convenience of deriving the variational lower bound. Under the type-II MABN, we need to compute $\mathbb{E}_{q(\mb{A})}[\log Z_i]$ for all $i$, where $Z_i=1/C_{p}(\kappa\|\sum_{j=1}^{i-1}\tilde{\mb{a}}_j\|_2)$ is the partition function of $p(\mb{\tilde{a}}_i|\textrm{pa}(\mb{\tilde{a}}_i))$. The analytical form of this expectation is difficult to derive as well due to the complexity of the $C_p(x)$ function: $C_p(x)=\frac{x^{p/2-1}}{(2\pi)^{p/2}I_{p/2-1}(x)}$ where $I_{p/2-1}(x)$ is the modified Bessel function of the first kind at order $p/2-1$. To address this issue, we derive an upper bound of $\log Z_i$ and compute the expectation of the upper bound, which is relatively easy to do. Consequently, we obtain a further lower bound of the variational lower bound and learn the variational and model parameters w.r.t the new lower bound. 
 
Now we proceed to derive the upper bound of $\log Z_i$, which equals to $\log \int \exp(\kappa (-\sum_{j=1}^{i-1}\mb{\tilde{a}}_j)\cdot \mb{\tilde{a}}_i) \mathrm{d}\mb{\tilde{a}}_i$.
Applying the inequality $\log \int\exp(x)\mathrm{d}x\leq \gamma+\int\log(1+\exp(x-\gamma))\mathrm{d}x$ \citep{bouchard2007efficient}, where $\gamma$ is a variational parameter, we have
\begin{equation}
\log Z_i
\leq \gamma+\int \log(1+\exp(\kappa (-\sum_{j=1}^{i-1}\mb{\tilde{a}}_j)\cdot \mb{\tilde{a}}_i-\gamma) \mathrm{d}\mb{\tilde{a}}_i.
\end{equation}
Then applying the inequality $\log(1+e^{-x})\leq \log(1+e^{-\xi})-\frac{x-\xi}{2}-\frac{1/2-g(\xi)}{2\xi}(x^2-\xi^2)$ \citep{bouchard2007efficient}, where $\xi$ is another variational parameter and $g(\xi)=1/(1+\exp(-\xi))$, we have
\begin{equation}
\log Z_i
\leq \gamma+\int [\log(1+e^{-\xi})-\frac{\kappa (\sum\limits_{j=1}^{i-1}\mb{\tilde{a}}_j)\cdot \mb{\tilde{a}}_i+\gamma-\xi}{2}-\frac{1/2-g(\xi)}{2\xi}((\kappa (\sum\limits_{j=1}^{i-1}\mb{\tilde{a}}_j)\cdot \mb{\tilde{a}}_i+\gamma)^2-\xi^2)] \mathrm{d}\mb{\tilde{a}}_i . \end{equation}
Finally, applying the following integrals on a high-dimensional sphere: (1) $\int_{\|\mb{y}\|_2=1} 1\mathrm{d}\mb{y}=\frac{2\pi^{(p+1)/2}}{\Gamma(\frac{p+1}{2})}$, (2) $\int_{\|\mb{y}\|_2=1} \mb{x}^\top\mb{y}\mathrm{d}\mb{y}=0$, (3) $\int_{\|\mb{y}\|_2=1} (\mb{x}^\top\mb{y})^2\mathrm{d}\mb{y}\leq \|\mb{x}\|_2^2\frac{2\pi^{(p+1)/2}}{\Gamma(\frac{p+1}{2})}$, we get
\begin{equation}
\label{eq:logz}
\begin{array}{l}
\log Z_i\leq-\frac{1/2-g(\xi)}{2\xi}\kappa^2\|\sum\limits_{j=1}^{i-1}\mb{\tilde{a}}_j\|_2^2\frac{2\pi^{(p+1)/2}}{\Gamma(\frac{p+1}{2})}+\gamma+[\log(1+e^{-\xi})+\frac{\xi-\gamma}{2}+\frac{1/2-g(\xi)}{2\xi}(\xi^2-\gamma^2)]\frac{2\pi^{(p+1)/2}}{\Gamma(\frac{p+1}{2})}\\
\end{array}
\end{equation}
The expectation of this upper bound is much easier to compute.
Specifically, we need to tackle $\mathbb{E}_{q(\mb{A})}[\|\sum_{j=1}^{i-1}\mb{\tilde{a}}_j\|_2^2]$, which can be computed as
\begin{eqnarray}
\mathbb{E}_{q(\mb{A})}[\|\sum_{j=1}^{i-1}\mb{\tilde{a}}_j\|_2^2]
&=&\mathbb{E}_{q(\mb{A})}[\sum_{j=1}^{i-1}\mb{\tilde{a}}_j^\top \mb{\tilde{a}}_j+
\sum_{j=1}^{i-1}\sum_{k\neq j}^{i-1}\mb{\tilde{a}}_j^\top \mb{\tilde{a}}_k] \nonumber \\
&=&\sum\limits_{j=1}^{i-1}\mathrm{tr}(\mathbb{E}_{q(\mb{\tilde{a}}_j)}[\mb{\tilde{a}}_j\mb{\tilde{a}}_j^\top])+
\sum\limits_{j=1}^{i-1}\sum\limits_{k\neq j}^{i-1}\mathbb{E}_{q(\mb{\tilde{a}}_j)}[\mb{\tilde{a}}_j]^\top \mathbb{E}_{q(\mb{\tilde{a}}_k)}[\mb{\tilde{a}}_k] \nonumber  \\
&=&\sum\limits_{j=1}^{i-1}\mathrm{tr}(\mathrm{cov}(\mb{\tilde{a}}_j))+
\sum\limits_{j=1}^{i-1}\sum\limits_{k=1}^{i-1}\mathbb{E}_{q(\mb{\tilde{a}}_j)}[\mb{\tilde{a}}_j]^\top \mathbb{E}_{q(\mb{\tilde{a}}_k)}[\mb{\tilde{a}}_k],
\end{eqnarray}
where $\mathbb{E}_{q(\mb{\tilde{a}}_j)}[\mb{\tilde{a}}_j]=A_p(\hat{\kappa})\hat{\mb{a}}_j$, $\mathrm{cov}(\mb{\tilde{a}}_j)=\frac{h(\hat{\kappa})}{\hat{\kappa}}\mathbf{I}+(1-2\frac{\nu+1}{\hat{\kappa}}h(\hat{\kappa})-h^2(\hat{\kappa}))\hat{\mb{a}}_j \hat{\mb{a}}_j^\top$, $h(\hat{\kappa})=\frac{I_{\nu+1}(\hat{\kappa})}{I_{\nu}(\hat{\kappa})}$, $A_p(\hat{\kappa})=\frac{I_{p/2}(\hat{\kappa})}{I_{p/2-1}(\hat{\kappa})}$ and $\nu=p/2-1$.

\paragraph{MCMC Sampling}
One potential drawback of the variational inference approach is that a large approximation error can be incurred if the variational distribution is far from the true posterior. We further present an alternative approximation inference method --- Markov chain Monte Carlo (MCMC) \citep{gilks2005markov}, which draws samples from the $\textit{exact}$ posterior distribution and uses the samples to represent the posterior. Specifically we choose the Metropolis-Hastings (MH) algorithm \citep{gilks2005markov} which generates samples from an adaptive proposal distribution, computes acceptance probabilities based on the unnormalized true posterior and uses the acceptance probabilities to decide whether a sample should be accepted or rejected. The most commonly used proposal distribution is based on random walk: the newly proposed sample $t+1$ comes from a random perturbation around the previous sample $t$. For the directional variables $\{\mb{\tilde{a}}_i\}_{i=1}^{K}$ and magnitude variables $\{g_i\}_{i=1}^{K}$, we define the proposal distributions to be a von Mises-Fisher distribution and a normal distribution respectively:
\begin{equation}
\begin{array}{l}
q(\mb{\tilde{a}}_i^{(t+1)}|\mb{\tilde{a}}_i^{(t)})=
C_{p}(\hat{\kappa})\exp\left(\hat{\kappa}\mb{\tilde{a}}_i^{(t+1)}\cdot\mb{\tilde{a}}_i^{(t)}\right)\\
q(g_i^{(t+1)}|g_i^{(t)})=\frac{1}{\sigma\sqrt{2\pi}}\exp\left\{-\frac{( g_i^{(t+1)}-g_i^{(t)})^2}{2\sigma^2}\right\}.
\end{array}
\end{equation}
$g_i^{(t+1)}$ is required to be positive, but the Gaussian distribution may generate non-positive samples. To address this problem, we
adopt a truncated sampler \citep{Wilkinson} which repeatedly
draws samples until a positive value
is obtained. Under such a truncated sampling scheme, the
MH acceptance ratio needs to be modified accordingly. Please refer to \citep{Wilkinson} for details. 
 
MH eventually converges to a stationary distribution where the generated samples represent the true posterior. The downside of MCMC is that it could take a long time to converge, which is usually computationally less efficient than variational inference \citep{hoffman2013stochastic}.
Under the MH algorithm, the MABN prior facilitates better efficiency compared with the DPP prior. In each iteration, the MABN prior needs to be evaluated, whose complexity is quadratic in the component number $K$ whereas evaluating the DPP has a cubic complexity in $K$.

\subsection{Diversity-Promoting Posterior Regularization}
 
In practice, one may desire to achieve more than one diversity-promoting effects in LVMs. For example, the mutual angular regularizer \citep{xie2015diversifying} aims to encourage the pairwise angles between components to have not only large mean, but also small variance such that the components are uniformly ``different'' from each other and evenly spread out to different directions in the space. It would be extremely difficult, if ever possible, to define a proper prior that can accommodate all desired effects. For instance, the MABN priors defined above can encourage the mutual angles to have large mean, but are unable to promote small variance.
To overcome such inflexibility of the prior control method, we resort to a \textit{posterior regularization} approach \citep{zhu2014bayesian}. Instead of designing a Bayesian prior to encode the diversification desideratum and indirectly influencing the posterior, posterior regularization directly imposes a control over the post-data distributions to achieve certain goals.
Giving prior $\pi(\mb{A})$ and data likelihood $p(\mathcal{D}|\mb{A})$, computing the posterior $p(\mb{A}|\mathcal{D})$ is equivalent to solving the following optimization problem \citep{zhu2014bayesian}
\begin{equation}
\label{eq:pr}
\textrm{sup}_{q(\mb{A})}\quad\mathbb{E}_{q(\mb{A})}[\log p(\mathcal{D}|\mb{A})\pi(\mb{A})]-\mathbb{E}_{q(\mb{A})}[\log q(\mb{A})],
\end{equation}
where $q(\mb{A})$ is any valid probability distribution.
The basic idea of posterior regularization is to impose a certain regularizer $\mathcal{R}(q(\mb{A}))$ over $q(\mb{A})$ to incorporate prior knowledge and structural bias \citep{zhu2014bayesian} and solve the following regularized problem
\begin{equation}
\label{eq:postreg}
\textrm{sup}_{q(\mb{A})}\quad\mathbb{E}_{q(\mb{A})}[\log p(\mathcal{D}|\mb{A})\pi(\mb{A})]-\mathbb{E}_{q(\mb{A})}[\log q(\mb{A})]+\lambda\mathcal{R}(q(\mb{A})),
\end{equation}
where $\lambda$ is a tradeoff parameter. Through properly designing $\mathcal{R}(q(\mb{A}))$, many diversity-promoting effects can be flexibly incorporated. Here we present a specific example while noting that many other choices are applicable. Gaining insight from \citep{xie2015diversifying}, we define $\mathcal{R}(q(\mb{A}))$ as
\begin{equation}
\label{eq:mar}
\begin{array}{l}
\Omega(\{\mathbb{E}_{q(\mb{a}_i)}[\mb{a}_i]\}_{i=1}^K)=\frac{1}{K(K-1)}\sum_{i=1}^{K}\limits\sum_{j\neq i}^{K}\limits\theta_{ij}
-\gamma\frac{1}{K(K-1)}\sum\limits_{i=1}^{K}\sum\limits_{j\neq i}^{K}(\theta_{ij}-\frac{1}{K(K-1)}\sum\limits_{p=1}^{K}\sum\limits_{q\neq p}^{K}\theta_{pq})^{2} ,
\end{array}
\end{equation}
where $\theta_{ij}=\textrm{arccos}(\frac{|\mathbb{E}[\mb{a}_i]\cdot\mathbb{E}[\mb{a}_j]|}{\|\mathbb{E}[\mb{a}_i]\|_2\|\mathbb{E}[\mb{a}_j]\|_2})$ is the non-obtuse angle measuring the dissimilarity between $\mathbb{E}[\mb{a}_i]$ and $\mathbb{E}[\mb{a}_j]$, and the regularizer is defined as the mean of pairwise angles minus their variance. The intuition behind this regularizer is: if the mean of angles is larger (indicating these vectors are more different from each other on the whole) and the variance of the angles is smaller (indicating these vectors evenly spread out to different directions), then these vectors are more diverse. Note that it is very difficult to design priors to simultaneously achieve these two effects.

While posterior regularization is more flexible, it lacks some strengths possessed by the prior control method for our consideration of diversifying latent variable models. First, prior control is a more natural way of incorporating prior knowledge, with solid theoretical foundation. Second, prior control can facilitate sampling based algorithms that are not applicable for the above posterior regularization.\footnote{Note that it does exist some examples of posterior regularization that have nice sampling-based algorithms, such as the max-margin topic models with a Gibbs classifier~\citep{zhu2014gibbs}.}
In sum, the two approaches have complementary advantages and should be chosen according to specific problem context.
 
\begin{figure}
\begin{center}
\includegraphics[width=0.4\columnwidth]{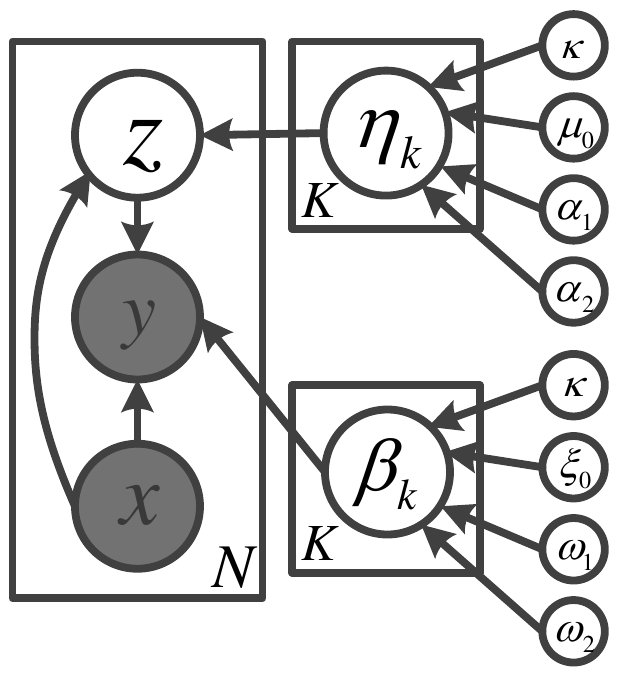}
\end{center}
\caption{Bayesian Mixture of Experts with Mutual Angular Prior}
\label{fig:me}
\end{figure}
\subsection{``Diversifying'' Bayesian Mixture of Experts Model}
In this section, we apply the two approaches developed above to ``diversify'' the Bayesian mixture of experts model (BMEM) \citep{waterhouse1996bayesian}.
\subsubsection{BMEM with Mutual Angular Prior}

The mixture of experts model (MEM) \citep{jordan1994hierarchical} has been widely used for machine learning tasks where the distribution of input data is so complicated that a single model (``expert'') cannot be effective for all the data. MEM assumes that the input data is inherently belonging to multiple latent groups and one single ``expert'' is allocated to each group to handle the data therein. Here we consider a classification task whose goal is to learn binary linear classifiers given the training data $\mathcal{D}=\{(\mb{x}_i,y_i)\}_{i=1}^{N}$, where $\mb{x}$ is the input feature vector and $y_i\in\{1,0\}$ is the class label. We assume there are $K$ latent experts where each expert is a classifier with coefficient vector $\bs\beta$. Given a test example $\mb{x}$, it first goes through a gate function that decides which expert is best suitable to classify this example and the decision is made in a probabilistic way. A discrete variable $z$ is utilized to indicate the selected expert and the probability that $z=k$ (assigning example $\mb{x}$ to expert $k$) is $\frac{\exp(\bs\eta_k^\top \mb{x})}{\sum_{j=1}^K\exp(\bs\eta_j^\top \mb{x})}$, where $\bs\eta_k$ is a coefficient vector characterizing the selection of expert $k$. Given the selected expert, the example is classified using the coefficient vector $\bs\beta$ corresponding to that expert. As described in Figure \ref{fig:me}, the generative process of $\{(\mb{x}_i,y_i)\}_{i=1}^{N}$ is as follows
\begin{itemize}
\item For $i=1,\cdots,N$
\begin{itemize}
\setlength\itemsep{0.1em}
\item Draw $z_i\sim \text{Multi}(\bs\zeta)$, where $\zeta_k=\frac{\exp(\bs\eta_k^\top \mb{x}_i)}{\sum_{j=1}^K\exp(\bs\eta_j^\top \mb{x}_i)}$
\item Draw $y_i\sim \text{Bernoulli}(\frac{1}{1+\exp(-\bs\beta_{z_i}^\top \mb{x}_i)})$.
\end{itemize}
\end{itemize}
 
As of now, the model parameters $\mb{B}=\{\bs\beta_k\}_{k=1}^{K}$ and $\mb{H}=\{\bs\eta_k\}_{k=1}^{K}$ are deterministic variables. Next we place a prior over them to enable Bayesian learning \citep{waterhouse1996bayesian} and desire this prior to be able to promote diversity among the experts to retain the advantages of ``diversifying'' LVMs as stated before. The mutual angular Bayesian network prior can be applied to achieve this goal
\begin{equation*}
p(\mb{B})=C_{p}(\kappa)\exp\left(\kappa \mu_{0}^\top \mb{\tilde{\bs\beta}}_1\right)\prod_{i=2}^{K}C_{p}(\kappa)\exp\left(\kappa(-\frac{\sum_{j=1}^{i-1}\tilde{\bs\beta}_j}{||\sum_{j=1}^{i-1}\mb{\tilde{\bs\beta}}_j||_2})^\top \mb{\tilde{\bs\beta}}_i\right)
\prod_{i=1}^{K}\frac{\alpha_2^{\alpha_1}g_i^{\alpha_1-1}e^{-g_i\alpha_2}}{\Gamma(\alpha_1)},
\end{equation*}
\begin{equation*}
p(\mb{H})=C_{p}(\kappa)\exp\left(\kappa \xi_{0}^\top \mb{\tilde{\bs\eta}}_1\right)\prod_{i=2}^{K}C_{p}(\kappa)\exp\left(\kappa(-\frac{\sum_{j=1}^{i-1}\tilde{\bs\eta}_j}{||\sum_{j=1}^{i-1}\mb{\tilde{\bs\eta}}_j||_2})^\top \mb{\tilde{\bs\eta}}_i\right)
\prod_{i=1}^{K}\frac{\omega_2^{\omega_1}h_i^{\omega_1-1}e^{-h_i\omega_2}}{\Gamma(\omega_1)},
\end{equation*}
where $\bs\beta_k=g_k\tilde{\bs\beta}_k$ and $\bs\eta_k=h_k\tilde{\bs\eta}_k$.

\subsubsection{BMEM with Mutual Angular Posterior Regularization}
As an alternative approach, the diversity in BMEM can be imposed by placing the mutual angular regularizer (Eq.(\ref{eq:mar})) over the post-data posteriors \citep{zhu2014bayesian}. Here we instantiate the general diversity-promoting posterior regularization defined in Eq.(\ref{eq:postreg}) to BMEM, by specifying the following parametrization. The latent variables in BMEM include $\mb{B}$, $\mb{H}$ and $\mb{z}=\{z_i\}_{i=1}^{N}$ and the post-data distribution over them is defined as $q(\mb{B},\mb{H},\mb{z})=q(\mb{B})q(\mb{H})q(\mb{z})$. For computational tractability, we define $q(\mb{B})$ and $q(\mb{H})$ to be: $q(\mb{B})=\prod_{k=1}^{K}q(\tilde{\bs\beta}_k)q(g_k)$ and $q(\mb{H})=\prod_{k=1}^{K}q(\tilde{\bs\eta}_k)q(h_k)$ where $q(\tilde{\bs\beta}_k)$, $q(\tilde{\bs\eta}_k)$ are von Mises-Fisher distributions and $q(g_k)$, $q(h_k)$ are gamma distributions, and define $q(\mb{z})$ to be multinomial distributions: $q(\mb{z})=\prod_{i=1}^{N}q(z_i|\bs\phi_i)$ where $\bs\phi_i$ is a multinomial vector. The priors over $\mb{B}$ and $\mb{H}$ are specified to be: $\pi(\mb{B})=\prod_{k=1}^{K}p(\tilde{\bs\beta}_k)p(g_k)$ and $\pi(\mb{H})=\prod_{k=1}^{K}p(\tilde{\bs\eta}_k)p(h_k)$ where $p(\tilde{\bs\beta}_k)$, $p(\tilde{\bs\eta}_k)$ are vMF distributions and $p(g_k)$, $p(h_k)$ are gamma distributions.
Under such parametrization, we solve the following diversity-promoting posterior regularization problem
\begin{equation}
\begin{array}{ll}
\textrm{sup}_{q(\mb{B},\mb{H},\mb{z})}&\quad\mathbb{E}_{q(\mb{B},\mb{H},\mb{z})}[\log p(\{y_i\}_{i=1}^{N},\mb{z}|\mb{B},\mb{H})\pi(\mb{B},\mb{H})]-\mathbb{E}_{q(\mb{B},\mb{H},\mb{z})}[\log q(\mb{B},\mb{H},\mb{z})]\\
&\quad+\lambda_1\Omega(\{\mathbb{E}_{q(\tilde{\bs\beta}_k)}[\tilde{\bs\beta}_k]\}_{k=1}^K)+\lambda_2\Omega(\{\mathbb{E}_{q(\tilde{\bs\eta}_k)}[\tilde{\bs\eta}_k]\}_{k=1}^K).
\end{array}
\end{equation}
Note that other parametrizations are also valid, such as placing Gaussian priors over $\mb{B}$ and $\mb{H}$ and setting $q(\mb{B})$, $q(\mb{H})$ to be Gaussian.

\section{Diversity-Promoting Bayesian Nonparametric Modeling}
In the last section, we study how to promote diversity among a finite number of components in parametric LVMs. In this section, we investigate how to achieve this goal in nonparametric LVMs where the component number is infinite in principle. We extend the mutual angular Bayesian network (MABN) prior defined in last section to an Infinite Mutual Angular (IMA) prior that encourages infinitely many components to have large angles. In this prior, the components are mutually dependent, which incurs great challenges for posterior inference. We develop an efficient sampling algorithm based on slice sampling~\citep{tehstick} and Riemann manifold Hamiltonian Monte Carlo~\citep{girolami2011riemann}. We apply the IMA prior to induce diversity in the infinite latent feature model (ILFM) \citep{ghahramani2005infinite}.
 
\subsection{Bayesian Nonparametric Latent Variable Models}
 
A BN-LVM consists of an infinite number of components, each parameterized by a vector. For example, in Dirichlet process Gaussian mixture model (DP-GMM) \citep{rasmussen1999infinite,blei2006variational}, the components are \textit{clusters}, each parameterized with a Gaussian mean vector. In Indian buffet process latent feature model (IBP-LFM) \citep{ghahramani2005infinite}, the components are \textit{features}, each parameterized by a weight vector. Given these infinitely many components, BN-LVMs design some proper mechanism to select one or a finite subset of them to model each observed data example. For example, in DP-GMM, a Chinese restaurant process (CRP) \citep{aldous1985exchangeability} is designed to assign each data example to one of the infinite number of clusters. In IBP-LFM, an Indian buffet process (IBP) \citep{ghahramani2005infinite} is utilized to select a finite set of features from the infinite feature pool to reconstruct each data example. A BN-LVM typically consists of two priors. One is a base distribution from which the parameter vectors of components are drawn. The other is a stochastic process -- such as CRP and IBP -- which designates how to select components to model data. The prior studied in this paper belongs to the first regime. It is commonly assumed that parameter vectors of the components are \textit{independently} drawn from the same base distribution. For example, in both DP-GMM and IBP-LFM, the mean vectors and weight vectors are independently drawn from a Gaussian distribution. In this paper, we aim to design a prior that encourages the component vectors to be mutually different and ``diverse", under which the component vectors are not independent any more, which presents great challenges for posterior inference.
 
\subsection{Infinite Mutual Angular Prior} 
In the MABN prior, the components are added one by one. Each new component is encouraged to have large angles with previous ones. This adding process can be repeated infinitely many times, resulting in a prior that encourages an infinite number of components to have large mutual angles
\begin{equation}
    p(\{\mb{\widehat{w}}_i\}_{i=1}^{\infty})=p(\mb{\widehat{w}}_1)\prod_{i=2}^{\infty}p(\mb{\widehat{w}}_i|\textrm{pa}(\mb{\widehat{w}}_i))
\end{equation}
The factorization theorem \citep{koller2009probabilistic} of Bayesian network ensures that $p(\{\mb{\widehat{w}}_i\}_{i=1}^{\infty})$ integrates to one. The magnitudes $\{r_i\}_{i=1}^{\infty}$ do not affect angles (hence diversity), which can be generated independently from a gamma distribution.
 
To this end, the generative process of $\{\mb{w}_i\}_{i=1}^{\infty}$ can be summarized as follows:
\begin{itemize}
\setlength\itemsep{0em}
\item Sample $\mb{\widehat{w}}_1\sim \textrm{vMF}(\bs\mu_{0},\kappa)$
\item For $i=2,\cdots,\infty$, sample $\mb{\widehat{w}}_i\sim \textrm{vMF}(-\frac{\sum_{j=1}^{i-1}\widehat{\mb{w}}_j}{\|\sum_{j=1}^{i-1}\widehat{\mb{w}}_j\|_2},\kappa)$
\item For $i=1,\cdots,\infty$,
sample $r_i\sim \textrm{Gamma}(\alpha_1,\alpha_2)$
\item For $i=1,\cdots,\infty$, $\mb{w}_i=\mb{\widehat{w}}_i r_i$
\end{itemize}
The probability distribution over $\{\mb{w}_i\}_{i=1}^{\infty}$ can be written as
\begin{equation}
\label{eq:map_1_np}
\begin{array}{l}
p(\{\mb{w}_i\}_{i=1}^{\infty})=C_{p}(\kappa)\exp(\kappa \mu_{0}^\top\mb{\widehat{w}}_1)\prod_{i=2}^{\infty}C_{p}(\kappa)
\exp(\kappa 
(-\frac{\sum_{j=1}^{i-1}\widehat{\mb{w}}_j}{\|\sum_{j=1}^{i-1}\widehat{\mb{w}}_j\|_2})^\top\widehat{\mb{w}}_i)\prod_{i=1}^{\infty}\frac{\alpha_2^{\alpha_1}r_i^{\alpha_1-1} e^{-r_i\alpha_2}}{\Gamma(\alpha_1)}
\end{array}
\end{equation}
 
\subsection{Diversity-Promoting Infinite Latent Feature Model}
In this section, using infinite latent feature model (ILFM) \citep{griffiths2005infinite} as an instance of BN-LVM, we showcase how to promote diversity among the components therein with the IMA prior. Given a set of data examples $\mc{X}=\{\mb{x}_n\}_{n=1}^{N}$ where $\mb{x}_n\in \mathbb{R}^{D}$, ILFM aims to invoke a finite subset of features from an infinite feature collection $\mathcal{W}=\{\mb{w}_k\}_{k=1}^{\infty}$ to construct these data examples. Each feature (which is a component in this LVM) is parameterized by a vector $\mb{w}_k\in \mathbb{R}^{D}$. For each data example $\mb{x}_n$, a subset of features are selected to construct it. The selection is denoted by a binary vector $\mb{z}_n\in \{0,1\}^\infty$ where $z_{nk}=1$ denotes the $k$-th feature is invoked to construct the $n$-th example and $z_{nk}=0$ otherwise. Given the parameter vectors of features $\{\mb{w}_k\}_{k=1}^{\infty}$ and the selection vector $\mb{z}_n$, the example $\mb{x}_n$ can be represented as: $\mb{x}_n\sim \mathcal{N}(\sum_{k=1}^{\infty}z_{nk}\mb{w}_k,\sigma^2\mb{I})$. The binary selection vectors $\mathcal{Z}=\{\mb{z}_n\}_{n=1}^{N}$ can be either drawn from an Indian buffet process (IBP)~\citep{ghahramani2005infinite} or a stick-breaking construction~\citep{tehstick}. Let $\mu_k$ be the prior probability that feature $k$ is present in a data example and the features are permuted such that their prior probabilities are in a decreasing ordering: $\mu_{(1)}>\mu_{(2)}>\cdots$. According to the stick-breaking construction, these prior probabilities can be generated in the following way: $\nu_{k}\sim \text{Beta}(\alpha,1)$, $\mu_{(k)}=\nu_{k}\mu_{(k-1)}=\prod_{l=1}^{k}\nu_{l}$. Given $\mu_{(k)}$, the binary indicator $z_{nk}$ is generated as $z_{nk}|\mu_{(k)}\sim \text{Bernoulli}(\mu_{(k)})$. To reduce the redundancy among the features, we impose the IMA prior over their parameter vectors $\mathcal{W}$ to encourage them to be mutually different, which results in an IMA-LFM model.

\subsection{Algorithm}
In this section, we develop a sampling algorithm to infer the posteriors of $\mathcal{W}$ and $\mathcal{Z}$ in the IMA-LFM model. Two major challenges need to be addressed. First, the prior over $\mathcal{W}$ is not conjugate to the likelihood function $p(\mb{x})$. Second, the parameter vectors $\mathcal{W}$ are usually of high-dimensional, rendering slow mixing. To address the first challenge, we adopt the slicing sampling algorithm \citep{tehstick}. This algorithm introduces an auxiliary slice variable $
s|\mc{Z},\mu_{(1:\infty)} \sim \text{Uniform}[0,\mu^*]$, where $\mu^*=\text{min}\{1, \underset{k:\exists n, z_{nk}=1}{\textrm{min}}\mu_k\}$ is the prior probability of the last active feature. A feature $k$ is active if there exists an example $n$ such that $z_{nk}=1$ and is inactive otherwise. In the sequel, we discuss the sampling of other variables.

\paragraph{Sample New Features} Let $K^*$ be the maximal feature index with $\mu_{(K^*)}>s$ and $K^+$ be the index such that all active features have index $k<K^+$ ($K^+$ itself would be inactive feature). If the new value of $s$ makes $K^*\geq K^+$, then we draw $K^*-K^++1$ new (inactive) features, including the parameter vectors and prior probabilities. The prior probabilities $\{\mu_{(k)}\}$ are drawn sequentially from $p(\mu_{(k)}|\mu_{(k-1)}) \propto \exp(\alpha \sum_{n=1}^{N}\frac{1}{n}(1-\mu_{(k)})^n)\mu_{(k)}^{\alpha-1}(1-\mu_{(k)})^N \mathbb{I}(0\leq \mu_{(k)}\leq \mu_{(k-1)})$ using adaptive rejection sampling (ARS) \citep{gilks1992adaptive}. The parameter vectors are drawn sequentially from
\begin{equation}
\begin{array}{l}
p(\mb{w}_k|\{\mb{w}_j\}_{j=1}^{k-1})=
p(\mb{\widehat{w}}_k|\{\mb{\widehat{w}}_j\}_{j=1}^{k-1})p(r_k)
=C_{p}(\kappa)\exp(\kappa(-\frac{\sum_{j=1}^{k-1}\widehat{\mb{w}}_j}{\|\sum_{j=1}^{k-1}\widehat{\mb{w}}_j\|_2})^\top\widehat{\mb{w}}_k) \frac{\alpha_2^{\alpha_1}r_i^{\alpha_1-1} e^{-r_i\alpha_2}}{\Gamma(\alpha_1)}
\end{array}
\end{equation}
where we draw $\mb{\widehat{w}}_k$ from $p(\mb{\widehat{w}}_k|\{\mb{\widehat{w}}_j\}_{j=1}^{k-1})$ which is a von Mises-Fisher distribution and draw $r_k$ from a Gamma distribution, then multiply $\mb{\widehat{w}}_k$ and $r_k$ together since they are independent. For each new feature $k$, the corresponding binary selection variables $z_{:,k}$ are initialized to zero.

\paragraph{Sample Existing $\mu_{(k)}$ $(1\leq k\leq K^+-1)$} We sample $\mu_{(k)}$ from $p(\mu_{(k)}|\text{rest})
\propto \mu_{(k)}^{m_{k}-1}(1-\mu_{(k)})^{N-m_{k}}\mathbb{I}(\mu_{(k+1)}\leq \mu_{(k)}\leq \mu_{(k-1)})$, where $m_{k}=\sum_{n=1}^{N}z_{nk}$.

\paragraph{Sample $z_{nk}$ $(1\leq n\leq N, 1\leq k\leq K^*)$}
Given $s$, we only need to sample $z_{nk}$ for $k\leq K^*$ from $p(z_{nk}=1|\text{rest})\propto \frac{\mu_{(k)}}{\mu^*}p(\mb{x}_n|z_{n,\neg k},z_{nk}=1,\{\mb{w}_j\}_{j=1}^{K^+})$, where $p(\mb{x}_n|z_{n,\neg k},z_{nk}=1,\{\mb{w}_j\}_{j=1}^{K^+})
=\mathcal{N}(\mb{x}_n|\mb{w}_k+\sum_{j\neq k}^{K^+}z_{nj}\mb{w}_{j},\sigma\mb{I})$ and $z_{n,\neg k}$ denotes all other elements in $\mb{z}$
except the $k$-th one and $\mb{w}_{j}=\mb{\widehat{w}}_j r_j$.

\paragraph{Sample $\mb{w}_{k}$ $(k=1,\cdots,K^+)$} We draw $\mb{w}_{k}=\widetilde{\mb{w}}_{k}r_k$ from the following conditional probability
\begin{equation}
\label{eq:update_a}
\begin{array}{l}
p(\widetilde{\mb{w}}_{k}r_k|\text{rest})
\propto p(\widetilde{\mb{w}}_{k}r_k|\{\mb{w}_j\}_{j\neq k}^{K^+})\prod\limits_{n=1}^{N}p(\mb{x}_n|z_{n,1:K^+},\{\mb{w}_j\}_{j\neq k}^{K^+},\widetilde{\mb{w}}_{k}r_k)
\end{array}
\end{equation}
where $p(\widetilde{\mb{w}}_{k}r_k|\{\mb{w}_j\}_{j\neq k}^{K^+})\propto p(\widetilde{\mb{w}}_{k}r_k|\{\mb{w}_i\}_{i=1}^{k-1})
\prod_{j=k+1}^{K^+}p(\mb{w}_j|\{\mb{w}_i\}_{i\neq k}^{j-1},\widetilde{\mb{w}}_{k}r_k)$ and $p(\mb{x}_n|z_{n,1:K^+}$ $,\{\mb{w}_j\} _{j\neq k}^{K^+}$,  $\widetilde{\mb{w}}_{k}r_k)=\mathcal{N}(\mb{x}_n|\widetilde{\mb{w}}_{k}r_k+\sum_{j\neq k}^{K^+}z_{nj}\mb{w}_{j},\sigma\mb{I})$. 
In the vanilla IBP latent feature model \citep{ghahramani2005infinite}, the prior over $\mb{w}_k$ is a Gaussian distribution, which is conjugate to the Gaussian likelihood function. In that case, the posterior $p(\mb{w}_{k}|\text{rest})$ is again a Gaussian, from which samples can be easily drawn. But in Eq.(\ref{eq:update_a}), the posterior does not have a closed form expression since the prior $p(\widetilde{\mb{w}}_{k}r_k|\{\mb{w}_j\}_{j\neq k}^{K^+})$ is no longer a conjugate prior, making the sampling very challenging.

We sample $\widetilde{\mb{w}}_{k}$ and $r_k$ separately. $r_k$ can be efficiently sampled using the Metropolis-Hastings (MH) \citep{hastings1970monte} algorithm. For $\widetilde{\mb{w}}_{k}$ which is a random vector, the sampling is much more difficult. The random walk based MH algorithm suffers slow mixing when the dimension of $\widetilde{\mb{w}}_{k}$ is large (which is typically the case in LVMs). In addition, $\widetilde{\mb{w}}_{k}$ lies on a unit sphere. 
The sampling algorithm should preserve this geometric constraint. To address these two issues, we study a Riemann manifold Hamiltonian Monte Carlo (RM-HMC) method \citep{girolami2011riemann,byrne2013geodesic}. HMC leverages the Hamiltonian dynamics to produce distant proposals for the Metropolis-Hastings algorithm, enabling a faster exploration of the state space and faster mixing. The RM-HMC algorithm introduces an auxiliary vector $\mb{v}\in\mathbb{R}^{d}$ and defines a Hamiltonian function $H(\mb{\widehat{w}}_{k},\mb{v})=-\log p(\mb{\widehat{w}}_{k}|\text{rest}) +\log|G(\mb{\widehat{w}}_{k})|+\frac{1}{2}\mb{v}^\top G(\mb{\widehat{w}}_{k})^{-1} \mb{v}$, where $G$ is the metric tensor associated with the Riemann manifold, which in our case is a unit sphere. After a transformation of coordinate system, $H(\mb{\widehat{w}}_{k},\mb{v})$ can be re-written as
\begin{equation}
H(\mb{\widehat{w}}_{k},\mb{v})=-\log p(\mb{\widehat{w}}|rest)+\frac{1}{2}\mb{v}^\top \mb{v}
\end{equation}
$p(\mb{\widehat{w}}_{k}|\text{rest})$ needs not to be normalized and $\log p(\mb{\widehat{w}}_{k}|\text{rest})\propto \kappa(-\frac{\sum_{j=1}^{k-1}\widehat{\mb{w}}_j}{\|\sum_{j=1}^{k-1}\widehat{\mb{w}}_j\|_2})^\top\widehat{\mb{w}}_k
+\sum_{j=k+1}^{K^+}\\\kappa(-\frac{\sum_{i\neq k}^{j-1}\widehat{\mb{w}}_i+\mb{\widehat{w}}_{k}r_k}{\|\sum_{i\neq k}^{j-1}\widehat{\mb{w}}_i+\mb{\widehat{w}}_{k}r_k\|_2})^\top\widehat{\mb{w}}_j +\sum_{n=1}^{N}\frac{1}{\sigma}(\mb{x}_n-\sum_{j\neq k}^{K^+}z_{nj}\mb{w}_j)^\top \mb{\widehat{w}}_{k}r_k-\frac{1}{2\sigma}\|\mb{\widehat{w}}_{k}r_k\|_2^2$. A new sample of $\mb{\widehat{w}}_{k}$ can be generated by approximately solving a system of differential equations characterizing the Hamiltonian dynamics on the manifold \citep{girolami2011riemann}.
\begin{figure}
\begin{algorithm}[H]
\caption{A Manifold Hamiltonian Monte Carlo Algorithm for Sampling $\mb{\widehat{w}}$}\label{alg:euclid}
\begin{algorithmic}
\STATE 01. $\mb{v}\sim\mathcal{N}(0,\mb{I})$
\STATE 02. $\mb{v}\gets \mb{v}-(\mb{I}-\mb{\widehat{w}} \mb{\widehat{w}}^\top)\mb{v}$
\STATE 03. $h\gets \log p(\mb{\widehat{w}}|rest)-\frac{1}{2}\mb{v}^\top \mb{v}$
\STATE 04. $\mb{\widehat{w}}^*\gets \mb{\widehat{w}}$
\STATE 05. \textbf{for} {$\tau=1,\cdots, T$} \textbf{do}
\STATE 06.\quad$\mb{v}\gets \mb{v}+\frac{\epsilon}{2}\nabla_{\mb{\widehat{w}}^*}\log p(\mb{\widehat{w}}^*|rest)$
\STATE 07.\quad $\mb{v}\gets \mb{v}-\mb{\widehat{w}} \mb{\widehat{w}}^\top\mb{v}$
\STATE 08.\quad $\mb{\widehat{w}}^*\gets \cos(\epsilon\|\mb{v}\|_2)\mb{\widehat{w}}^*+\|\mb{v}\|_2^{-1}\sin(\epsilon\|\mb{v}\|_2)\mb{v}$
\STATE 09.\quad $\mb{v}\gets -\|\mb{v}\|_2\sin(\epsilon\|\mb{v}\|_2)\mb{\widehat{w}}^*+\cos(\epsilon\|\mb{v}\|_2)\mb{v}$
\STATE 10.\quad $\mb{v}\gets \mb{v}+\frac{\epsilon}{2}\nabla_{\mb{\widehat{w}}^*}\log p(\mb{\widehat{w}}^*|rest)$
\STATE 11.\quad $\mb{v}\gets \mb{v}-\mb{\widehat{w}} \mb{\widehat{w}}^\top\mb{v}$
\STATE 12. \textbf{end for}
\STATE 13. $h^*\gets \log p(\mb{\widehat{w}}^*|rest)-\frac{1}{2}\mb{v}^\top \mb{v}$
\STATE 14. $u\sim \text{uniform}(0,1)$
\STATE 15. \textbf{if} {$u<\text{exp}(h^*-h)$}
	\STATE 16.\quad  $\mb{\widehat{w}}\gets \mb{\widehat{w}}^*$
\STATE 17. \textbf{end if}
\end{algorithmic}
\label{alg:mhmc}
\end{algorithm}
\end{figure}
Following \citep{byrne2013geodesic}, we solve this problem based upon geodesic flow, which is shown in Line 6-11 in Algorithm \ref{alg:mhmc}. Line 6 performs an update of $\mb{v}$ according to the Hamiltonian dynamics, where $\nabla_{\mb{\widehat{w}}^*}\log p(\mb{\widehat{w}}^*|rest)$ denotes the gradient of $\log p(\mb{\widehat{w}}^*|rest)$ w.r.t $\mb{\widehat{w}}^*$. Line 7 performs the transformation of coordinate system. Line 8-9 calculate the geodesic flow on unit sphere. Line 10-11 repeat the update of $\mb{v}$ as done in Line 6-7. These procedures are repeated $T$ times to generate a new sample $\mb{\widehat{w}}^*$, which then goes through an acceptance/rejection procedure (Line 3, 13-17).

\section{Experiments}
 
We now present experimental results for both the parametric and nonparametric latent variable models (LVMs) to demonstrate the effectiveness on encouraging diversity.
 
\subsection{Parametric LVM}
We first present the results with parametric LVMs. Using Bayesian mixture of experts model as an instance, we conducted experiments to verify the effectiveness and efficiency of the two proposed approaches.
\label{sec:exp}
\paragraph{Datasets}
We used two binary-classification datasets. The first one is the Adult-9 \citep{platt1999fast} dataset, which has $\sim$33K training instances and $\sim$16K testing instances. The feature dimension is 123. The other dataset is SUN-Building compiled from the SUN \citep{xiao2010sun} dataset, which contains $\sim$6K building images and 7K non-building images randomly sampled from other categories, where 70\% of images are used for training and the rest for testing. We use SIFT \citep{lowe1999object} based bag-of-words to represent the images with a dimensionality of 1000.
\paragraph{Experimental Settings}
To understand the effects of diversification in Bayesian learning, we compare the following methods: (1) mixture of experts model (MEM) with L2 regularization (MEM-L2) where the L2 regularizer is imposed over ``experts'' independently; (2) MEM with mutual angular regularization \citep{xie2015diversifying} (MEM-MAR) where the ``experts'' are encouraged to be diverse; (3) Bayesian MEM with a Gaussian prior (BMEM-G) where the ``experts'' are independently drawn from a Gaussian prior; (4) BMEM with mutual angular Bayesian network priors (type I or II) where the MABN favors diverse ``experts'' (BMEM-MABN-I, BMEM-MABN-II); BMEM-MABN-I is inferred with MCMC sampling and BMEM-MABN-II is inferred with variational inference; (5) BMEM with posterior regularization (BMEM-PR).

The key parameters involved in the above methods are: (1) the regularization parameter $\lambda$ in MEM-L2, MEM-MAR, BMEM-PR; (2) the concentration parameter $\kappa$ in the mutual angular priors in BMEM-MABN-(I,II); (3) the concentration parameter $\hat{\kappa}$ in the variational distribution in BMEM-MABN-II.
All parameters were tuned using 5-fold cross validation. Besides internal comparison, we also compared with 5 baseline methods, which are among the most widely used classification approaches that achieve the state of the art performance. They are: (1) kernel support vector machine (KSVM) \citep{burges1998tutorial}; (2) random forest (RF) \citep{breiman2001random}; (3) deep neural network (DNN) \citep{hinton2006reducing}; (4) Infinite SVM (ISVM) \citep{zhu2011infinite}; (5) BMEM with DPP prior (BMEM-DPP) \citep{kulesza2012determinantal}, in which a Metropolis-Hastings sampling algorithm was adopted\footnote{Variational inference and Gibbs sampling \citep{affandi2013approximate} are both not applicable.}. The kernels in KSVM and BMEM-DPP are both radial basis function kernel. Parameters of the baselines were tuned using 5-fold cross validation.
\paragraph{Results}

\begin{table}[t]
\centering
\begin{tabular}{|c|c|c|c|c|c|}
\hline
K & 5&10&20&30\\
\hline
MEM-L2&82.6&83.8&84.3&84.7\\
\hline
MEM-MAR &85.3&86.4&86.6&87.1\\
\hline
BMEM-G&83.4&84.2&84.9&84.9\\
\hline
BMEM-MABN-I &\tb{87.1}&\tb{88.3}&88.6&\tb{88.9}\\
\hline
BMEM-MABN-II&86.4&87.8&88.1&88.4\\
\hline
BMEM-PR&86.2&87.9&\tb{88.7}&88.1\\
\hline
\end{tabular}
\caption{Classification accuracy (\%) on Adult-9 dataset}\label{table:retrieval_20news}
\end{table}

\begin{table}[t]
\centering
\begin{tabular}{|c|c|c|c|c|}
\hline
K & 5&10&20&30\\
\hline
MEM-L2 &76.2&78.8&79.4&79.7\\
\hline
MEM-MAR &81.3&82.1&82.7&82.3\\
\hline
BMEM-G&76.5&78.6&80.2&80.4\\
\hline
BMEM-MABN-I &\tb{82.1}&83.6&\tb{85.3}&\tb{85.2}\\
\hline
BMEM-MABN-II&80.9&82.8&84.9&84.1\\
\hline
BMEM-PR&81.7&\tb{84.1}&83.8&84.9\\
\hline
\end{tabular}
\caption{Classification accuracy (\%) on SUN-Building dataset}\label{table:retrieval_15scenes}
\end{table}

\begin{table*}[t]
\small
\centering
\begin{tabular}{|c|c|c|c|c|c|c|c|c|c|}
\hline
Category ID& C18& C17& C12 &C14 &C22 &C34 &C23 &C32 &C16 \\
\hline
Num. of Docs & 5281 & 4125 & 1194 & 741 & 611& 483 & 262 &208 &192\\
\hline
BMEM-G Accuracy (\%)&87.3 	& 88.5&	75.7&	70.1	&71.6	&64.2&55.9&	57.4	&51.3
\\
\hline
BMEM-MABN-I Accuracy (\%) & 88.1&86.9		&74.7&72.2	&70.5	&	67.3&68.9	&70.1	&	65.5
\\
\hline
Relative Improvement (\%)&1.0&-1.8&-1.3&2.9&-1.6&4.6&18.9&18.1&21.7	
\\
\hline
\end{tabular}
\caption{Accuracy on 9 subcategories of the CCAT category in the RCV1.Binary dataset}
\label{tb:prec_cat_rcv1}
\end{table*}

\begin{table}[t]
\centering
\begin{tabular}{|c|c|c|}
\hline
& Adult-9&SUN-Building\\
\hline
KSVM&85.2&84.2\\
RF&87.7&85.1\\
DNN&87.1&84.8\\
ISVM&85.8&82.3\\
BMEM-DPP&86.5&84.5\\
\hline
BMEM-MABN-I&\tb{88.9}&\tb{85.3}\\
BMEM-MABN-II&88.4&84.9\\
BMEM-PR&88.7&84.9\\
\hline
\end{tabular}
\caption{Classification accuracy (\%) on two datasets}
\label{table:cmp_ap}
\end{table}

Table \ref{table:retrieval_20news} and \ref{table:retrieval_15scenes} show the classification accuracy under different numbers of ``experts'' on the Adult-9 and SUN-Building datasets respectively. From these two tables, we observe that: (1) diversification can greatly improve the performance of Bayesian MEM, which can be seen from the comparison between diversified BMEM methods and their non-diversified counterparts, such as BMEM-MABN-(I,II) versus BMEM-G, and BMEM-PR versus BMEM-G. (2) Bayesian learning achieves better performance than point estimation, which is manifested by comparing BMEM-G with MEM-L2, and BMEM-MABN-(I,II)/BMEM-PR with MEM-MAR. (3) BMEM-MAR-I works better than BMEM-MABN-II and BMEM-PR. The reason is that BMEM-MAR-I inferred with MCMC draws samples from the \textit{exact} posterior while BMEM-MABN-II and BMEM-PR inferred with variational inference seek an \textit{approximation} of the posterior.

\begin{table}[t]
\centering
\begin{tabular}{|c|c|c|}
\hline
& Adult-9&SUN-Building\\
\hline
BMEM-DPP&8.2&11.7\\
BMEM-MABN-I&7.5&10.5\\
BMEM-MABN-II&2.9&4.1\\
BMEM-PR&3.3&4.9\\
\hline
\end{tabular}
\caption{Training time (hours) of different methods with $K=30$}
\label{table:runtime}
\end{table}
 
Recall that the goals of promoting diversity in LVMs are to reduce model size without sacrificing modeling power and effectively capture infrequent patterns. Here we empirically verify whether these two goals can be achieved.
Regarding the first goal, we compare diversified BMEM methods BMEM-MABN-(I,II)/BMEM-PR with non-diversified counterpart BMEM-G and check whether diversified methods with a small number of components $K$ which entails low computational complexity can achieve performance as good as non-diversified methods with large $K$.
It can be observed that BMEM-MABN-(I,II)/BMEM-PR with a small $K$ can achieve accuracy that is comparable to or even better than BMEM-G with a large $K$. For example, on the Adult-9 dataset (Table \ref{table:retrieval_20news}), with 5 experts BMEM-MABN-I is able to achieve an accuracy of $87.1\%$, which cannot be achieved by BMEM-G with even 30 experts.
This corroborates the effectiveness of diversification in reducing model size (hence computational complexity) without compromising performance.

To verify the second goal -- capturing infrequent patterns, from the RCV1 \citep{lewis2004rcv1} dataset we pick up a subset of documents (for binary classification) such that the popularity of categories (patterns) is in power-law distribution. Specifically, we choose documents from 9 subcategories (the 1st row of Table \ref{tb:prec_cat_rcv1}) of the CCAT category as the positive instances, and randomly sample 15K documents from non-CCAT categories as negative instances.
The 2nd row shows the number of documents in each of the 9 categories. The distribution of document frequency is in a power-law fashion, where frequent categories (such as C18 and C17) have a lot of documents while infrequent categories (such as C32 and C16) have a small amount of documents. The 3rd and 4th row show the accuracy achieved by BMEM-G and BMEM-MABN-I on each category. The 5th row shows the relative improvement of BMEM-MABN-I over BMEM-G, which is defined as $\frac{A_{bmem\_mabn}-A_{bmem\_g}}{A_{bmem\_g}}$, where $A_{bmem\_mabn}$ and $A_{bmem\_g}$ denote the accuracy achieved by BMEM-MABN-I and BMEM-G respectively. While achieving accuracy comparable to BMEM-G over the frequent categories, BMEM-MABN-I obtains much better performance on infrequent categories. For example, the relative improvements on infrequent categories C32 and C16 are 18.1\% and 21.7\%. This demonstrates that BMEM-MABN-I can effectively capture the infrequent patterns.
 
Table \ref{table:cmp_ap} presents the comparison with the state of the art classification methods. As one can see, our method achieves the best performances on both datasets. In particular, BMEM-MAR-(I,II) work better than BMEM-DPP, demonstrating the proposed mutual angular priors possess ability that is better than or comparable to the DPP prior in inducing diversity.
 
Table \ref{table:runtime} compares the time (hours) taken by each method to achieve convergence, with $K$ set to 30. BMEM-MABN-II inferred with variational inference (VI) is more efficient than BMEM-MABN-I inferred with MCMC sampling, due to the higher efficiency of VI than MCMC. BMEM-PR is solved with an optimization algorithm which is more efficient than the sampling algorithm in BMEM-MABN-I. BMEM-MABN-II and BMEM-PR are more efficient than BMEM-DPP where DPP inhibits the adoption of VI.

\begin{table*}[t]
\scriptsize
\centering
\begin{tabular}{|c|c|c|c|c|}
\hline
Dataset&\#Examples&Dimension &\#Classes &Description\\
\hline
Yale&722&1032&-&faces images\\
\hline
Block-Images&1000&36&-&noisy overlays of four binary shapes on a grid\\
\hline
AR&2600&1598&-&faces with lighting, accessories\\
\hline
EEG&4400&32&-&EEG recording on various tasks\\
\hline
Piano&10000&161&-&DFT of a piano recording\\
\hline
Reuters &7195&5000&9&Reuters news articles\\
\hline
TDT&9394&5000&30&Nist Topic Detection and Tracking (TDT) corpus\\
\hline
20-News &18846&5000&20&documents from 20 newsgroups\\
\hline
15-Scenes &4485&1000&15& images from 15 scene categories\\
\hline
Caltech-101 &9144&1000&101&images from 101 object categories\\
\hline
\end{tabular}
\caption{Statistics of Datasets}
\label{tb:data_stat}
\end{table*}

\subsection{Nonparametric LVM}
We evaluate the effectiveness of the IMA prior in alleviate overfitting, reducing model size without sacrificing modeling power and capturing infrequent patterns, on a wide range of datasets.
\paragraph{Datasets}
We used ten datasets from different domains including texts, images, sound and EEG signals. Their statistics are summarized in Table \ref{tb:data_stat}. The first five datasets are represented as raw data without feature extraction. The documents in Reuters\footnote{\url{http://www.daviddlewis.com/resources/testcollections/reuters21578/}}, TDT\footnote{\url{http://www.itl.nist.gov/iad/mig/tests/tdt/2004/}} and 20-News\footnote{\url{http://qwone.com/~jason/20Newsgroups/}} are represented with bag-of-words vectors, weighted using tf-idf. The images in 15-Scenes \citep{lazebnik2006beyond} and Caltech-101 \citep{fei2007learning} are represented with visual bag-of-words vectors based on SIFT \citep{lowe2004distinctive} features. The train/test split of each dataset is 70\%/30\%. The results are averaged over five random splits. 
\paragraph{Experimental Setup}
For each dataset, we use IMA-LFM to learn the latent features $\mathcal{W}$ on the training set, then use $\mathcal{W}$ to reconstruct the test data. The reconstruction performance is measured with L2 error (the smaller, the better) and log-likelihood (the larger, the better). Meanwhile, we use $\mathcal{W}$ to infer the representations $\mathcal{Z}$ of test data and perform data clustering on $\mathcal{Z}$. Clustering performance is measured using accuracy and normalized mutual information (NMI) (the higher, the better) \citep{cai2005document}. We compared with two baselines: Indian buffet process LFM (IBP-LTM) \citep{griffiths2005infinite} and Pitman-Yor process LFM (PYP-LFM) \citep{teh2009indian}. Following \citep{doshi2009accelerated}, all datasets are centered to have zero mean. $\kappa$ is set to 1. $\sigma^2$ is set to $0.25\hat{\sigma}$ where $\hat{\sigma}$ is the standard deviation of data across all dimensions. $\alpha$ is set to 2.  
 
\begin{table}
\centering
\begin{tabular}{|c|c|c|c|}
\hline
Dataset&IBP-LFM&PYP-LFM&IMA-LFM\\
\hline
Yale&447$\pm$7&432$\pm$3&\tb{419}$\pm$4\\
\hline
Block ($\times 10^{-2}$)&6.3$\pm$0.4 &5.8$\pm$0.1&\tb{4.4}$\pm$0.2\\
\hline
AR&926$\pm$4&939$\pm$11&\tb{871}$\pm$7\\
\hline
EEG (+$2.1 \times 10^{6}$) &5382$\pm$34&3731$\pm$15&\tb{575}$\pm$21\\
\hline
Piano ($\times 10^{-4}$)&5.3$\pm$0.1&5.7$\pm$0.2&\tb{4.2}$\pm$0.2\\
\hline
\end{tabular}
\caption{L2 Test Error}
\label{tb:l2}
\end{table}
\paragraph{Results}
Table \ref{tb:l2} and \ref{tb:llh} present the L2 error and likelihood (mean$\pm$standard error) on the test set of the first five datasets. As can be seen, IMA-LFM achieves much lower L2 error and higher likelihood than IBP-LFM and PYP-LFM. Table \ref{tb:txt_clus_ac} and \ref{tb:txt_clus_nmi} show the clustering accuracy and NMI (mean$\pm$standard error) on the last 5 datasets which have class labels. IMA-LFM outperforms the two baseline methods with a large margin. We conjecture the reasons are two-fold. First, IMA places a diversity-biased structure over the latent features, which alleviates overfitting. In both IBP-LFM and PYP-LFM, the weight vectors of latent features are drawn independently from a Gaussian distribution, which is unable to characterize the relations among features. In contrast, IMA-LFM imposes a structure over the features, encouraging them to be ``diverse" and less redundant. This structural constraint reduces model complexity of LFM, thus alleviating overfitting on the training data and achieving better reconstruction of the test data. Second, ``diversified" features presumably have higher representational power and are able to capture richer information and subtle aspects of data, thus achieving a better modeling effect.

\begin{table}
\centering
\begin{tabular}{|c|c|c|c|}
\hline
Dataset&IBP-LFM&PYP-LFM& IMA-LFM\\
\hline
Yale&-16.4$\pm$0.3&-14.9 $\pm$0.4&\tb{-12.7}$\pm$0.1\\
\hline
Block-Image&-2.1$\pm$0.2&-1.8$\pm$0.1 &\tb{-1.4}$\pm$0.1\\
\hline
AR&-13.9$\pm$0.3& -14.6$\pm$0.7&\tb{-8.5}$\pm$0.4\\
\hline
EEG&-14133$\pm$54&-12893 $\pm$73 &\tb{-9735}$\pm$32\\
\hline
Piano&-6.8$\pm$0.6&-6.9$\pm$0.2& \tb{-4.2}$\pm$0.5\\
\hline
\end{tabular}
\caption{Test Likelihood}
\label{tb:llh}
\end{table}

Table \ref{tb:num_feats} shows the number of features (mean$\pm$standard error) obtained by each model when the algorithm converges. Analyzing Table \ref{tb:l2}-\ref{tb:num_feats} simultaneously, we see that IMA-LFM uses much fewer features to achieve better performance than the baselines. For instance, on the Reuters dataset, with 294 features, IMA-LFM achieves a 48.2\% clustering accuracy. In contrast, IBP-LFM uses 60 more features, but achieves 2.8\% lower accuracy. This suggests that IMA is able to reduce the size of LFM (the number of features) without sacrificing the modeling power. 
Because of IMA's diversity-promoting mechanism, the learned features bear less redundancy and are highly complementary to each other. Each feature captures a significant amount of information. As a result, a small number of such features are sufficient to model data well. In contrast, the features in IBP-LFM and PYP-LFM are drawn independently from a base distribution, which lacks the mechanism to reduce redundancy. IMA achieves more significant reduction of feature number on datasets with larger dimensions. This is possibly because higher-dimensional data contains more redundancy, giving IMA a larger room to improve.
 
\begin{table}
\centering
\begin{tabular}{|c|c|c|c|}
\hline
& IBP-LFM&PYP-LFM&IMA-LFM\\
\hline
Reuters&45.4$\pm$0.3&43.1$\pm$0.4&\tb{48.2}$\pm$0.6\\
\hline
TDT&48.3$\pm$0.7&47.5$\pm$0.3&\tb{53.2}$\pm$0.4\\
\hline
20-News&21.5$\pm$0.1&23.7$\pm$0.2&\tb{25.2}$\pm$0.1\\
\hline
15-Scenes&22.7$\pm$0.2&21.9$\pm$0.4&\tb{25.3}$\pm$0.2\\
\hline
Caltech-101&11.6$\pm$0.4&12.1$\pm$0.1&\tb{14.7}$\pm$0.2\\
\hline
\end{tabular}
\caption{Clustering Accuracy (\%)}
\label{tb:txt_clus_ac}
\end{table}

To verify whether IMA helps to better capture infrequent patterns, on the learned features of the Reuters dataset we perform a retrieval task and measure the precision@100 on each category. For each test document,
we retrieve 100 documents from the training set based on Euclidean distance.

Precision@100 is defined as $n$/100, where $n$ is the number
of retrieved documents that share the same category label with
the query document. We treat each category as a pattern and define its frequency as the number of documents belonging to it. 
A category with more than 1000 documents is labeled as frequent. Table \ref{tb:pc_prec} shows the per-category precision. The last row shows the relative improvement of IMA-LFM over IBP-LFM, defined as $(P_{imap}-P_{ibp})/P_{ibp}$.
As can be seen, on the infrequent categories 3-9, IMA-LFM achieves much better precision than IBP-LFM, while on the frequent categories 1 and 2, their performance are comparable. This demonstrates that IMA is able to better capture infrequent patterns without losing the modeling power on frequent patterns.

IMA promotes diversity among the components, which pushes some of them away from the frequent patterns toward infrequent patterns, giving the infrequent ones a better chance to be captured. On the 20-News dataset, we visualize the learned features. For a latent feature with parameter vector $\mb{w}$, we pick up the top 10 words corresponding to the largest values in $\mb{w}$. Table 8 shows 5 exemplar features learned by IBP-LFM and IMA-LFM. As can be seen, the features learned by IBP-LFM have much overlap and redundancy and are hard to distinguish, whereas those learned by IMA-LFM are more diverse.

\begin{table}
\centering
\begin{tabular}{|c|c|c|c|}
\hline
& IBP-LFM&PYP-LFM&IMA-LFM\\
\hline
Reuters&41.7$\pm$0.5&38.6$\pm$0.2&\tb{45.4}$\pm$0.4\\
\hline
TDT&44.2$\pm$0.1&46.7$\pm$0.3&\tb{49.6}$\pm$0.6\\
\hline
20-News&38.9$\pm$0.8&35.2$\pm$0.5&\tb{44.6}$\pm$0.9\\
\hline
15-Scenes&42.1$\pm$0.7&44.9$\pm$0.8&\tb{47.5}$\pm$0.2\\
\hline
Caltech-101&34.2$\pm$0.4&36.8$\pm$0.4&\tb{40.3}$\pm$0.3\\
\hline
\end{tabular}
\caption{Normalized Mutual Information (\%)}
\label{tb:txt_clus_nmi}
\end{table}

\begin{table}
\centering
\begin{tabular}{|c|c|c|c|}
\hline
& IBP-LFM&PYP-LFM&IMA-LFM\\
\hline
Yale&201$\pm$5& 220$\pm$8&\tb{165}$\pm$4\\
\hline
Block-Image&\tb{8}$\pm$2&9$\pm$4&11$\pm$4\\
\hline
AR&257$\pm$11&193$\pm$5  &\tb{176}$\pm$8 \\
\hline
EEG&14$\pm$2& \tb{9}$\pm$2&12$\pm$1\\
\hline
Piano&37$\pm$4&34$\pm$6&\tb{28}$\pm$3 \\
\hline
Reuters&354$\pm$12&326$\pm$5&\tb{294}$\pm$7\\
\hline      
TDT&297$\pm$6&311$\pm$9&\tb{274}$\pm$3\\        
\hline
20-News&442$\pm$8&408$\pm$3&\tb{369}$\pm$5\\
\hline
15-Scenes&192$\pm$3&218$\pm$5&\tb{171}$\pm$8\\        
\hline
Caltech-101&127$\pm$7&113$\pm$6&\tb{96}$\pm$6\\
\hline
\end{tabular}
\caption{Number of features}
\label{tb:num_feats}
\end{table}
\begin{table}[t]
\small
\centering
\begin{tabular}{|c|c|c|c|c|c|c|c|c|c|}
\hline
Category & 1& 2& 3& 4& 5& 6& 7& 8& 9\\
\hline
Frequency &3713& 2055& 321& 298& 245& 197& 142& 114& 110\\
\hline
IBP-LFM Precision@100 (\%)  &73.7 &45.1 &7.5 &8.3 &6.9 &7.2 &2.6& 3.8& 3.4\\
\hline
IMA-LFM Precision@100 (\%)  &81.5& 78.4& 36.2& 37.8& 29.1& 20.4& 8.3& 13.8& 11.6\\
\hline
Relative Improvement (\%)  &11 &74 &382 &355 & 321 &183 & 219 &263 & 241\\
\hline
\end{tabular}
\caption{Per-category precision@100 on the Reuters dataset}
\label{tb:pc_prec}
\end{table}
\begin{table*}[t]
\label{tb:20news_topics}
\scriptsize
\centering
\begin{tabular}{llll|llll}
\hline
\multicolumn{4}{c|}{IBP-LFM} & \multicolumn{4}{c}{IMA-LFM} \\
\hline
 Feature 1 &Feature 2  &Feature 3&Feature 4 
&Feature 1&Feature 2 &Feature 3&Feature 4 \\
\hline
government&saddam&nuclear&turkish&president&clinton&olympic&school\\
house&white&iraqi&soviet&clinton&government&team&great\\
baghdad&clinton&weapons&government&legal&lewinsky&hockey&institute\\
weapons&united&united&weapons&years&nuclear&good&program\\
tax&time&work&number&state&work&baseball&study\\
years&president&spkr&enemy&baghdad&minister&gold&japanese\\
white&baghdad&president&good&church&weapons&ball&office\\
united&iraq&people&don&white&india&medal&reading\\
state&un&baghdad&citizens&united&years&april&level\\
bill&lewinsky&state&due&iraqi&white&winter&number\\
\hline
\end{tabular}
\label{tb:vis}
\caption{Visualization of features learned on the 20-News dataset}
\end{table*}

\section{Conclusions}
We study how to promote diversity in Bayesian latent variable models, for the purpose of better capturing infrequent patterns and reducing model size without compromising modeling power. We define a mutual angular Bayesian network (MABN) prior that entails an inductive bias towards components having larger mutual angles and investigate a posterior regularization approach which directly applies regularizers over the post-data distributions to promote diversity. Approximate algorithms are developed for posterior inference under the MABN priors. With Bayesian mixture of experts model as a study case, empirical experiments demonstrate the effectiveness and efficiency of our methods.
 
We also study how to promote diversity among infinitely many components in Bayesian nonparametric latent variable models. We extend the MABN prior to an infinite mutual angular (IMA) prior that encourages an infinite number of components to have large mutual angles. We apply the IMA prior to the infinite latent feature model, to encourage the latent features therein to be diverse. An efficient posterior inference algorithm is developed. Experiments demonstrate that the IMA prior can effectively capture infrequent patterns, reduce model size without compromising modeling power and alleviate overfitting.

\newpage
 
\appendix
\section*{Appendix A. Variational Inference for LVMs with Type I MABN Prior}
In this section, we present details on how to derive the variational lower bound
\begin{equation}
\begin{array}{l}
\mathbb{E}_{q(\mb{A})}[\log p(\mathcal{D}|\mb{A})]+\mathbb{E}_{q(\mb{A})}[\log p(\mb{A})]-\mathbb{E}_{q(\mb{A})}[\log q(\mb{A})]
\end{array}
\end{equation}
where the variational distribution $q(\mb{A})$ is chosen to be
\begin{equation}
\begin{array}{l}
q(\mb{A})=
\prod_{k=1}^{K}q(\tilde{\mb{a}}_k)q(g_k)=\prod\limits_{k=1}^{K}
\textrm{vMF}(\tilde{\mb{a}}_k|\hat{\mb{a}}_k,\hat{\kappa})
\textrm{Gamma}(g_k|r_k,s_k)
\end{array}
\end{equation}
Among the three expectation terms, $\mathbb{E}_{q(\mb{A})}[\log p(\mb{A})]$ and $\mathbb{E}_{q(\mb{A})}[\log q(\mb{A})]$ are model-independent and we discuss how to compute them in this section. $\mathbb{E}_{q(\mb{A})}[\log p(\mathcal{D}|\mb{A})]$ depends on the specific LVM and a concrete example will be given in Appendix B.
 
First we introduce some equalities and inequalities used later on. Let $\mb{a}\sim \mathrm{vMF}(\bs\mu, \kappa)$, then\\
(\rom{1}) $\mathbb{E}[\mb{a}]=A_p(\kappa)\bs\mu$ where $A_p(\kappa)=\frac{I_{p/2}(\kappa)}{I_{p/2-1}(\kappa)}$, and $I_{v}(\cdot)$ denotes the modified Bessel function of the first kind at order $v$.\\
(\rom{2}) $\mathrm{cov}(\mb{a})=\frac{h(\kappa)}{\kappa}\mathbf{I}+(1-2\frac{\nu+1}{\kappa}h(\kappa)-h^2(\kappa))\bs\mu \bs\mu^{T}$, where $h(\kappa)=\frac{I_{\nu+1}(\kappa)}{I_{\nu}(\kappa)}$ and $\nu=p/2-1$. \\
Please refer to \citep{Abeywardana} for the derivation of $\mathbb{E}[\mb{a}]$ and $\mathrm{cov}(\mb{a})$.\\
(\rom{3}) $\mathbb{E}[\mb{a}^T\mb{a}]=\mathrm{tr}(\mathrm{cov}(\mb{a}))+A_p^2(\kappa)\bs\mu^T\bs\mu$.
\paragraph{Proof}
\begin{equation}
\begin{array}{l}
\mathbb{E}[\mathrm{tr}(\mb{a}^T\mb{a})]
=\mathbb{E}[\mathrm{tr}(\mb{a}\mb{a}^T)]=
\mathrm{tr}(\mathbb{E}[\mb{a}\mb{a}^T])\\
=\mathrm{tr}(\textrm{cov}(\mb{a})+\mathbb{E}[\mb{a}]\mathbb{E}[\mb{a}]^T)
=\mathrm{tr}(\textrm{cov}(\mb{a}))+\mathrm{tr}(\mathbb{E}[\mb{a}]\mathbb{E}[\mb{a}]^T)\\
=\mathrm{tr}(\textrm{cov}(\mb{a}))+A_p^2(\kappa)\bs\mu^T\bs\mu\\
\end{array}
\end{equation}
Let $g\sim \textrm{Gamma}(\alpha,\beta)$, then\\
(\rom{4}) $\mathbb{E}[g]=\frac{\alpha}{\beta}$\\
(\rom{5}) $\mathbb{E}[\log g]=\psi(\alpha)-\log\beta$\\
(\rom{6})
$\log \sum_{k=1}^K\exp(x_k)\leq \gamma+\sum_{k=1}^K\log(1+\exp(x_k-\gamma))$ and $\log \int\exp(x)\mathrm{d}x\leq \gamma+\int\log(1+\exp(x-\gamma))\mathrm{d}x$, where $\gamma$ is a variational parameter. See \citep{bouchard2007efficient} for the proof.\\
(\rom{7}) $\log(1+e^{-x})\leq \log(1+e^{-\xi})-\frac{x-\xi}{2}-\frac{1/2-g(\xi)}{2\xi}(x^2-\xi^2)$, $\log(1+e^{x})\leq \log(1+e^{\xi})+\frac{x-\xi}{2}-\frac{1/2-g(\xi)}{2\xi}(x^2-\xi^2)$,
where $\xi$ is a variational parameter and $g(\xi)=1/(1+\exp(-\xi))$. See \citep{bouchard2007efficient} for the proof.\\
(\rom{8}) $\int_{\|\mb{y}\|_2=1} 1\mathrm{d}\mb{y}=\frac{2\pi^{(p+1)/2}}{\Gamma(\frac{p+1}{2})}$, which is the surface area\footnote{\url{https://en.wikipedia.org/wiki/N-sphere}} of $p$-dimensional unit sphere. $\Gamma(\cdot)$ is the Gamma function.\\
(\rom{9})$\int_{\|\mb{y}\|_2=1} \mb{x}^T\mb{y}\mathrm{d}\mb{y}=0$, which can be shown according to the symmetry of unit sphere.\\
(\rom{10})$\int_{\|\mb{y}\|_2=1} (\mb{x}^T\mb{y})^2\mathrm{d}\mb{y}\leq \|\mb{x}\|_2^2\frac{2\pi^{(p+1)/2}}{\Gamma(\frac{p+1}{2})}$.
\paragraph{Proof}
\begin{equation}
\begin{array}{lll}
&&\int_{\|\mb{y}\|_2=1} (\mb{x}^T\mb{y})^2\mathrm{d}\mb{y}\\
&=&\|\mb{x}\|_2^2\int_{\|\mb{y}\|_2=1} ((\frac{\mb{x}}{\|\mb{x}\|_2})^T\mb{y})^2\mathrm{d}\mb{y}\\
&=&\|\mb{x}\|_2^2\int_{\|\mb{y}\|_2=1} (\mb{e}_1^T\mb{y})^2\mathrm{d}\mb{y}\\
&&(\text{according to the symmetry of unit sphere})\\
&\leq&\|\mb{x}\|_2^2\int_{\|\mb{y}\|_2=1} 1\mathrm{d}\mb{y}\\
&=&\|\mb{x}\|_2^2\frac{2\pi^{(p+1)/2}}{\Gamma(\frac{p+1}{2})}
\end{array}
\end{equation}
Given these equalities and inequalities, we can upper bound $\log Z_i$ (Eq.(\ref{eq:logz}) in Section~\ref{sec:vi}). Given this upper bound, we can derive a lower bound of $\mathbb{E}_{q(\mb{A})}[\log p(\mb{A})]$
\begin{equation}
\label{eq:lb}
\begin{array}{l}
\mathbb{E}_{q(\mb{A})}[\log p(\mb{A})]\\
=\mathbb{E}_{q(\mb{A})}[\log p(\mb{\tilde{a}}_1)\prod_{i=2}^{K}p(\mb{\tilde{a}}_i|\{\mb{\tilde{a}}_j\}_{j=1}^{i-1}) \prod_{i=1}^{K}q(g_i)]\\
=\mathbb{E}_{q(\mb{A})}[\log p(\mb{\tilde{a}}_1)\prod\limits_{i=2}^{K}\frac{\exp(\kappa (-\sum_{j=1}^{i-1}\mb{\tilde{a}}_j)\cdot \mb{\tilde{a}}_i)}{Z_i}\prod\limits_{i=1}^{K}\frac{\alpha_2^{\alpha_1}g_i^{\alpha_1-1} e^{-g_i\alpha_2}}{\Gamma(\alpha_1)}]\\
\geq \kappa \mu_{0}^{\mathsf{T}}\mathbb{E}_{q(\mb{\tilde{a}}_1)}[\mb{\tilde{a}}_1]+
\sum_{i=2}^{K} (-\kappa\sum_{j=1}^{i-1}\mathbb{E}_{q(\mb{\tilde{a}}_j)}[\mb{\tilde{a}}_j]\cdot \mathbb{E}_{q(\mb{\tilde{a}}_i)}[\mb{\tilde{a}}_i]
\quad-\gamma_i-(\log(1+e^{-\xi_i})+\frac{\xi_i-\gamma_i}{2}\\
\quad+\frac{1/2-g(\xi_i)}{2\xi_i}(\xi_i^2-\gamma_i^2))\frac{2\pi^{(p+1)/2}}{\Gamma(\frac{p+1}{2})}
+\frac{1/2-g(\xi_i)}{2\xi_i}\kappa^2\mathbb{E}_{q(\mb{A})}[\|\sum_{j=1}^{i-1}\mb{\tilde{a}}_j\|_2^2]\frac{2\pi^{(p+1)/2}}{\Gamma(\frac{p+1}{2})}
)+K(\alpha_1\log \alpha_2\\
\quad-\log\Gamma(\alpha_1))
+\sum\limits_{i=1}^{K}(\alpha_1-1)\mathbb{E}_{q(g_i)}[\log g_i
]-\alpha_2 \mathbb{E}_{q(g_i)}[g_i]+\textrm{const}\\
\geq \kappa A_p(\hat{\kappa}) \mu_{0}^{\mathsf{T}}\mb{\hat{a}}_1+
\sum_{i=2}^{K}
(-\kappa A_p(\hat{\kappa})^2\sum_{j=1}^{i-1}\mb{\hat{a}}_j\cdot \mb{\hat{a}}_i
-\gamma_i-(\log(1+e^{-\xi_i})+\frac{\xi_i-\gamma_i}{2}\\
\quad+\frac{1/2-g(\xi_i)}{2\xi_i}(\xi_i^2-\gamma_i^2))\frac{2\pi^{(p+1)/2}}{\Gamma(\frac{p+1}{2})}+\frac{1/2-g(\xi_i)}{2\xi_i}\kappa^2
(A_p^2(\hat{\kappa})\sum_{j=1}^{i-1}\sum_{k\neq j}^{i-1}\mb{\hat{a}}_j\cdot \mb{\hat{a}}_k+\sum_{j=1}^{i-1}(\mathrm{tr}(\Lambda_j)\\
\quad+A_p^2(\hat{\kappa})\mb{\hat{a}}_j^T\mb{\hat{a}}_j))
\frac{2\pi^{(p+1)/2}}{\Gamma(\frac{p+1}{2})}
)+K(\alpha_1\log \alpha_2-\log\Gamma(\alpha_1))+\sum_{i=1}^{K}(\alpha_1-1)(\psi(r_i)-\log(s_i))-\alpha_2 \frac{r_i}{s_i}+\textrm{const}\\
\end{array}
\end{equation}
where $\Lambda_j=\frac{h(\hat{\kappa})}{\hat{\kappa}}\mathbf{I}+(1-2\frac{\nu+1}{\hat{\kappa}}h(\hat{\kappa})-h^2(\hat{\kappa}))\mb{\hat{a}}_j \mb{\hat{a}}_j^{T}$.
 
The other expectation term $\mathbb{E}_{q(\mb{A})}[\log q(\mb{A})]$ can be computed as
\begin{equation}
\label{eq:entropy}
\begin{array}{l}
\mathbb{E}_{q(\mb{A})}[\log q(\mb{A})]\\
=\mathbb{E}_{q(\mb{A})}[\log \prod\limits_{k=1}^{K}
\textrm{vMF}(\tilde{\mb{a}}_k|\hat{\mb{a}}_k,\hat{\kappa})
\textrm{Gamma}(g_k|r_k,s_k)]\\
=\sum\limits_{k=1}^{K} \hat{\kappa}A_p(\hat{\kappa})\|\hat{\alpha}_k\|_2^2+r_k\log s_k-\log\Gamma(r_k)+(r_k-1)(\psi(r_k)-\log(s_k))-r_k
\end{array}
\end{equation}

\section*{Appendix B. VI for BMEM with Type I MABN}
In this section, we discuss how to derive the variational lower bound for BMEM with type I MABN. The latent variables are $\{\bs\beta_k\}_{k=1}^{K}$,$\{\bs\eta_k\}_{k=1}^{K}$,$ \{z_n\}_{n=1}^{N}$. The joint probability of all variables is
\begin{equation}
\begin{array}{l}
p(\{\bs\beta_k\}_{k=1}^{K},\{\bs\eta_k\}_{k=1}^{K}, \{\mb{x}_n,y_n, z_n\}_{n=1}^{N})\\
= p(\{y_n\}_{n=1}^N|\{\mb{x}_n\}_{n=1}^N,\{z_n\}_{n=1}^N,\{\bs\beta_k\}_{k=1}^{K})  p(\{z_n\}_{n=1}^N|\{\mb{x}_n\}_{n=1}^N,\{\bs\eta_k\}_{k=1}^{K})
p(\{\bs\beta_k\}_{k=1}^{K})p(\{\bs\eta_k\}_{k=1}^{K})
\end{array}
\end{equation}
To perform variational inference, we employ a mean field variational distribution
\begin{equation}
\begin{array}{l}
Q=q(\{\bs\beta_k\}_{k=1}^{K},\{\bs\eta_k\}_{k=1}^{K}, \{z_n\}_{n=1}^{N})\\
=\prod\limits_{k=1}^K q(\bs\beta_k) q(\bs\eta_k) \prod_{n=1}^N q(z_n)\\
=\prod\limits_{k=1}^{K}
\textrm{vMF}(\tilde{\bs\beta}_k|\hat{\bs\beta}_k,\hat{\kappa})
\textrm{Gamma}(g_k|r_k,s_k)
\textrm{vMF}(\tilde{\bs\eta}_k|\hat{\bs\eta}_k,\hat{\kappa})
\textrm{Gamma}(h_k|t_k,u_k) \prod\limits_{n=1}^{N}q(z_n|\bs\phi_n)
\end{array}
\end{equation}
Accordingly, the variational lower bound is
\begin{equation}
\begin{array}{l}
\mathbb{E}_{Q}[\log p(\{\bs\beta_k\}_{k=1}^{K},\{\bs\eta_k\}_{k=1}^{K}, \{\mb{x}_n,y_n, z_n\}_{n=1}^{N})]
-\mathbb{E}_{Q}[\log q(\{\bs\beta_k\}_{k=1}^{K},\{\bs\eta_k\}_{k=1}^{K}, \{z_n\}_{n=1}^{N})]\\
=\mathbb{E}_{Q}[\log p(\{y_n\}_{n=1}^N|\{\mb{x}_n\}_{n=1}^N,\{z_n\}_{n=1}^N,\{\bs\beta_k\}_{k=1}^{K})]
+ \mathbb{E}_{Q}[\log p(\{z_n\}_{n=1}^N|\{\mb{x}_n\}_{n=1}^N,\{\bs\eta_k\}_{k=1}^{K})]\\
+\mathbb{E}_{Q}[\log p(\{\bs\beta_k\}_{k=1}^{K})])
+\mathbb{E}_{Q}[\log p(\{\bs\eta_k\}_{k=1}^{K})])
-\mathbb{E}_{Q}[\log q(\{\bs\beta_k\}_{k=1}^{K})]-\mathbb{E}_{Q}[\log q(\{\bs\eta_k\}_{k=1}^{K})]\\
-\mathbb{E}_{Q}[\log q(\{z_n\}_{n=1}^{N})]\\
\end{array}
\end{equation}
where $\mathbb{E}_{Q}[\log p(\{\bs\beta_k\}_{k=1}^{K})]$ and $\mathbb{E}_{Q}[\log p(\{\bs\eta_k\}_{k=1}^{K})]$ can be lower bounded in a similar way as that in Eq.(\ref{eq:lb}). $\mathbb{E}_{Q}[\log q(\{\bs\beta_k\}_{k=1}^{K})]$ and $\mathbb{E}_{Q}[\log q(\{\bs\eta_k\}_{k=1}^{K})]$ can be computed in a similar manner as that in Eq.(\ref{eq:entropy}). Next we discuss how to compute the remaining expectation terms.

\subsection*{Compute $\mathbb{E}_{Q}[\log p(\{z_n\}_{n=1}^{N}|\{\bs\eta_k\}_{k=1}^{K},\{\mb{x}_n\}_{n=1}^{N})]$}
 
First, $p(\{z_n\}_{n=1}^{N}|\{\bs\eta_k\}_{k=1}^{K},\{\mb{x}_n\}_{n=1}^{N})$ is defined as
\begin{equation}
\begin{array}{l}
p(\{z_n\}_{n=1}^{N}|\{\bs\eta_k\}_{k=1}^{K},\{\mb{x}_n\}_{n=1}^{N})\\
=\prod\limits_{n=1}^{N}p(z_n|\mb{x}_n,\{\bs\eta_k\}_{k=1}^{K})\\
=\prod\limits_{n=1}^{N}\frac{\prod\limits_{k=1}^{K}[\exp(\bs\eta_k^{\mathsf{T}}\mb{x}_n)]^{z_{nk}}}{\sum_{j=1}^K\exp(\bs\eta_j^{\mathsf{T}}\mb{x}_n)}
\end{array}
\end{equation}
$\log p(\{z_n\}_{n=1}^{N}|\{\bs\eta_k\}_{k=1}^{K},\{\mb{x}_n\}_{n=1}^{N})$ can be lower bounded as
\begin{equation}
\begin{array}{l}
\log p(\{z_n\}_{n=1}^{N}|\{\bs\eta_k\}_{k=1}^{K},\{\mb{x}_n\}_{n=1}^{N})\\
=
\sum\limits_{n=1}^{N}\sum\limits_{k=1}^{K}z_{nk}\bs\eta_k^{\mathsf{T}}\mb{x}_n
-\log(\sum_{j=1}^K\exp(\bs\eta_j^{\mathsf{T}}\mb{x}_n))\\
=
\sum\limits_{n=1}^{N}\sum\limits_{k=1}^{K}z_{nk}h_k\tilde{\bs\eta}_k^{\mathsf{T}}\mb{x}_n
-\log(\sum_{j=1}^K\exp(\bs\eta_j^{\mathsf{T}}\mb{x}_n))\\
\geq
\sum\limits_{n=1}^{N}\sum\limits_{k=1}^{K}z_{nk}h_k\tilde{\bs\eta}_k^{\mathsf{T}}\mb{x}_n
-c_n-\sum_{j=1}^K
\quad \log(1+\exp(\bs\eta_j^{\mathsf{T}}\mb{x}_n-c_n)) \text{(Using Inequality VI)}\\
\geq
\sum\limits_{n=1}^{N}\sum\limits_{k=1}^{K}z_{nk}h_k\tilde{\bs\eta}_k^{\mathsf{T}}\mb{x}_n
-c_n
-\sum_{j=1}^K[\log(1+e^{-d_{nj}})-\frac{c_n-\bs\eta_j^{\mathsf{T}}\mb{x}_n-d_{nj}}{2}-\frac{1/2-g(d_{nj})}{2d_{nj}}((\bs\eta_j^{\mathsf{T}}\mb{x}_n-c_n)^2-d_{nj}^2)]\\
\quad\text{(Using Inequality VII)}\\
\end{array}
\end{equation}
The expectation of $\log p(\{z_n\}_{n=1}^{N}|\{\bs\eta_k\}_{k=1}^{K},\{\mb{x}_n\}_{n=1}^{N})$ can be lower bounded as
\begin{equation}
\begin{array}{l}
\mathbb{E}[\log p(\{z_n\}_{n=1}^{N}|\{\bs\eta_k\}_{k=1}^{K},\{\mb{x}_n\}_{n=1}^{N})]\\
=
A_p(\hat{\kappa})\sum\limits_{n=1}^{N}\sum\limits_{k=1}^{K}\phi_{nk}\frac{t_k}{u_k}\hat{\bs\eta}_k^{\mathsf{T}}\mb{x}_n
-c_n
-\sum_{j=1}^K[\log(1+e^{-d_{nj}})
-\frac{c_n-A_p(\hat{\kappa})\frac{t_j}{u_j}\hat{\bs\eta}_j^{\mathsf{T}}\mb{x}_n-d_{nj}}{2}\\
-\frac{1/2-g(d_{nj})}{2d_{nj}}(
\frac{t_j+t_j^2}{u_j^2}\mathbb{E}[\tilde{\bs\eta}_{j}^{\mathsf{T}}\mb{x}_n\mb{x}^{\mathsf{T}}_n\tilde{\bs\eta}_{j}]
-2c_nA_p(\hat{\kappa})\frac{t_j}{u_j}\hat{\bs\eta}_j^{\mathsf{T}}\mb{x}_n+c_n^2
-d_{nj}^2)]\\
\end{array}
\end{equation}
where
\begin{equation}
\label{eq:eta}
\begin{array}{l}
\mathbb{E}[\tilde{\bs\eta}_{k}^{\mathsf{T}}\mb{x}_n\mb{x}^{\mathsf{T}}_n\tilde{\bs\eta}_{k}]\\
=\mathbb{E}[\mathrm{tr}(\tilde{\bs\eta}_{k}^{\mathsf{T}}\mb{x}_n\mb{x}^{\mathsf{T}}_n\tilde{\bs\eta}_{k})]\\
=\mathbb{E}[\mathrm{tr}(\mb{x}_n\mb{x}^{\mathsf{T}}_n\tilde{\bs\eta}_{k}\tilde{\bs\eta}_{k}^{\mathsf{T}})]\\
=\mathrm{tr}(\mb{x}_n\mb{x}^{\mathsf{T}}_n\mathbb{E}[\tilde{\bs\eta}_{k}\tilde{\bs\eta}_{k}^{\mathsf{T}}])\\
=\mathrm{tr}(\mb{x}_n\mb{x}^{\mathsf{T}}_n(\mathbb{E}[\tilde{\bs\eta}_{k}]\mathbb{E}[\tilde{\bs\eta}_{k}]^T+\textrm{cov}(\tilde{\bs\eta}_{k})))\\
\end{array}
\end{equation}

\subsection*{Compute $\mathbb{E}[\log p(\{y_n\}_{n=1}^{N}|\{\bs\beta_k\}_{k=1}^{K},\{\mb{z}_n\}_{n=1}^{N})]$}
$p(\{y_n\}_{n=1}^{N}|\{\bs\beta_k\}_{k=1}^{K},\{\mb{z}_n\}_{n=1}^{N})$ is defined as
\begin{equation}
\begin{array}{l}
p(\{y_n\}_{n=1}^{N}|\{\bs\beta_k\}_{k=1}^{K},\{\mb{z}_n\}_{n=1}^{N})\\
=\prod\limits_{n=1}^{N}p(y_n|\mb{z}_n,\{\bs\beta_k\}_{k=1}^{K})\\
=\prod\limits_{n=1}^{N}\frac{1}{\prod\limits_{k=1}^{K}[1+\exp(-(2y_n-1)\bs\beta_{k}^{\mathsf{T}}\mb{x}_n)]^{z_{nk}}}
\end{array}
\end{equation}
$\log p(\{y_n\}_{n=1}^{N}|\{\bs\beta_k\}_{k=1}^{K},\{\mb{z}_n\}_{n=1}^{N})$ can be lower bounded by
\begin{equation}
\begin{array}{l}
\log p(\{y_n\}_{n=1}^{N}|\{\bs\beta_k\}_{k=1}^{K},\{\mb{z}_n\}_{n=1}^{N})
\\=-\sum\limits_{n=1}^{N}\sum\limits_{k=1}^{K}z_{nk}\log(1+\exp(-(2y_n-1)\bs\beta_{k}^{\mathsf{T}}\mb{x}_n))
\\\geq\sum\limits_{n=1}^{N}\sum\limits_{k=1}^{K}z_{nk}
[-\log(1+e^{-e_{nk}})+\frac{(2y_n-1)\bs\beta_{k}^{\mathsf{T}}\mb{x}_n-e_{nk}}{2}+\frac{1/2-g(e_{nk})}{2e_{nk}}((\bs\beta_{k}^{\mathsf{T}}\mb{x}_n)^2-e_{nk}^2)
]
\end{array}
\end{equation}
 
$\mathbb{E}[\log p(\{y_n\}_{n=1}^{N}|\{\bs\beta_k\}_{k=1}^{K},\{\mb{z}_n\}_{n=1}^{N})]$ can be lower bounded by
\begin{equation}
\begin{array}{l}
\mathbb{E}[\log p(\{y_n\}_{n=1}^{N}|\{\bs\beta_k\}_{k=1}^{K},\{\mb{z}_n\}_{n=1}^{N})]
\\\geq\sum\limits_{n=1}^{N}\sum\limits_{k=1}^{K}\phi_{nk}
[-\log(1+e^{-e_{nk}})+\frac{A_p(\hat{\kappa})\frac{r_k}{s_k}\hat{\bs\beta}_{k}^{\mathsf{T}}\mb{x}_n-e_{nk}}{2}+\frac{1/2-\sigma(e_{nk})}{2e_{nk}}(\frac{r_k+r_k^2}{s_k^2}\mathbb{E}[\tilde{\bs\beta}_{k}^{\mathsf{T}}\mb{x}_n\mb{x}^{\mathsf{T}}_n\tilde{\bs\beta}_{k}]-e^2_{nk})
]\\
\quad\text{(Using Inequality VII)}\\
\end{array}
\end{equation}
where $\mathbb{E}[\tilde{\bs\beta}_{k}^{\mathsf{T}}\mb{x}_n\mb{x}^{\mathsf{T}}_n\tilde{\bs\beta}_{k}]$ can be computed in a similar way to Eq.(\ref{eq:eta}).
 
\subsection*{Compute $\mathbb{E}[\log q(z_i)]$}
\begin{equation}
\mathbb{E}[\log q(z_i)]=\sum\limits_{k=1}^{K}\phi_{ik}\log \phi_{ik}
\end{equation}
 
In the end, we can get a lower bound of the variational lower bound, then learn all the parameters by optimizing the lower bound via coordinate ascent: In each iteration, we pick up a parameter $x$ and fix all other parameters, which leads to a sub-problem defined over $x$. Then we optimize the sub-problem w.r.t $x$. For some parameters, the optimal solution of the sub-problem is in closed form. If not the case, we optimize $x$ using gradient ascent method. This process iterates until convergence. We omit the detailed derivation here since it only involves basic algebra and calculus, which can be done straightforwardly.
 
\section*{Appendix C. Parameter Learning in the Metropolis-Hastings Algorithm}
The mutual angular Bayesian Network (MABN) prior is parameterized by several deterministic parameters including $\kappa$, $\bs\mu_0$, $\alpha_1$, $\alpha_2$. Among them, we tune $\kappa$ manually via cross validation and learn the others via an Expectation Maximization (EM) framework. Let $\mb{x}$ denote observed data, $\mb{z}$ denote all random variables and $\bs\theta$ denote deterministic parameters $\{\bs\mu_0,\alpha_1,\alpha_2\}$. EM is an algorithm aiming to learn $\bs\theta$ by maximize log-likelihood $p(\mb{x};\bs\theta)$ of data. It iteratively performs two steps until convergence. In the E step, the posterior $p(\mb{z}|\mb{x})$ is inferred with parameters $\bs\theta$ fixed. In the M step, $\bs\theta$ is learned by optimizing a lower bound of the log-likelihood $\mathbb{E}_{p(\mb{z}|\mb{x})}[\log p(\mb{x},\mb{z};\bs\theta)]$, where the expectation is computed w.r.t the posterior $p(\mb{z}|\mb{x})$ inferred in the E step. In our problem, we use the Metropolis-Hastings (MH) algorithm to infer the posterior $p(\mb{x};\bs\theta)$ at the E step, and learn parameters $\{\bs\mu_0,\alpha_1,\alpha_2\}$ at the M step. The parameters $\hat{\kappa}$ and $\sigma$ in proposal distributions are set manually.

\section*{Appendix D. Algorithm for Posterior Regularization of BMEM}
In this section, we present the algorithmic details of posterior regularization of BMEM. Recall the problem is
\begin{equation}
\label{eq:pr_obj1}
\begin{array}{ll}
\textrm{sup}_{q(\mb{B},\mb{H},\mb{z})}&\mathbb{E}_{q(\mb{B},\mb{H},\mb{z})}[\log p(\{y_i\}_{i=1}^{N},\mb{z}|\mb{B},\mb{H})\pi(\mb{B},\mb{H})]-\mathbb{E}_{q(\mb{B},\mb{H},\mb{z})}[\log q(\mb{B},\mb{H},\mb{z})]\\
&+\lambda_1\Omega(\{\mathbb{E}_{q(\tilde{\bs\beta}_k)}[\tilde{\bs\beta}_k]\}_{k=1}^K)+\lambda_2\Omega(\{\mathbb{E}_{q(\tilde{\bs\eta}_k)}[\tilde{\bs\eta}_k]\}_{k=1}^K)
\end{array}
\end{equation}
where $\mb{B}=\{\bs\beta_k\}_{k=1}^{K}$, $\mb{H}=\{\bs\eta_k\}_{k=1}^{K}$ and $\mb{z}=\{z_i\}_{i=1}^{N}$ are latent variables and the post-data distribution over them is defined as $q(\mb{B},\mb{H},\mb{z})=q(\mb{B})q(\mb{H})q(\mb{z})$. For computational tractability, we define $q(\mb{B})$ and $q(\mb{H})$ to be: $q(\mb{B})=\prod_{k=1}^{K}q(\tilde{\bs\beta}_k)q(g_k)$ and $q(\mb{H})=\prod_{k=1}^{K}q(\tilde{\bs\eta}_k)q(h_k)$ where $q(\tilde{\bs\beta}_k)$, $q(\tilde{\bs\eta}_k)$ are von-Mises Fisher distributions and $q(g_k)$, $q(h_k)$ are gamma distributions, and define $q(\mb{z})$ to be multinomial distributions: $q(\mb{z})=\prod_{i=1}^{N}q(z_i|\bs\phi_i)$ where $\bs\phi_i$ is a multinomial vector. The priors over $\mb{B}$ and $\mb{H}$ are specified to be: $\pi(\mb{B})=\prod_{k=1}^{K}p(\tilde{\bs\beta}_k)p(g_k)$ and $\pi(\mb{H})=\prod_{k=1}^{K}p(\tilde{\bs\eta}_k)p(h_k)$ where $p(\tilde{\bs\beta}_k)$, $p(\tilde{\bs\eta}_k)$ are von-Mises Fisher distributions and $p(g_k)$, $p(h_k)$ are gamma distributions.
 
The objective in Eq.(\ref{eq:pr_obj1}) can be further written as
\begin{equation}
\begin{array}{ll}
\mathbb{E}_{q(\mb{B},\mb{H},\mb{z})}[\log p(\{y_i\}_{i=1}^{N},\mb{z}|\mb{B},\mb{H})\pi(\mb{B},\mb{H})]
-\mathbb{E}_{q(\mb{B},\mb{H},\mb{z})}[\log q(\mb{B},\mb{H},\mb{z})]+\lambda_1\Omega(\{\mathbb{E}_{q(\tilde{\bs\beta}_k)}[\tilde{\bs\beta}_k]\}_{k=1}^K)\\
+\lambda_2\Omega(\{\mathbb{E}_{q(\tilde{\bs\eta}_k)}[\tilde{\bs\eta}_k]\}_{k=1}^K)\\
=\mathbb{E}_{q(\mb{B},\mb{z})}[\log p(\{y_i\}_{i=1}^{N}|\mb{z},\mb{B})]+\mathbb{E}_{q(\mb{H},\mb{z})}[\log p(\mb{z}|\mb{H})]
+\mathbb{E}_{q(\mb{H})}[\log \pi(\mb{H})]+\mathbb{E}_{q(\mb{B})}[\log \pi(\mb{B})]\\
-\mathbb{E}_{q(\mb{B})}[\log q(\mb{B})]
-\mathbb{E}_{q(\mb{H})}[\log q(\mb{H})]-\mathbb{E}_{q(\mb{z})}[\log q(\mb{z})]+\lambda_1\Omega(\{\mathbb{E}_{q(\tilde{\bs\beta}_k)}[\tilde{\bs\beta}_k]\}_{k=1}^K)
+\lambda_2\Omega(\{\mathbb{E}_{q(\tilde{\bs\eta}_k)}[\tilde{\bs\eta}_k]\}_{k=1}^K)
\end{array}
\end{equation}
 
Among these expectation terms, $\mathbb{E}_{q(\mb{B},\mb{z})}[\log p(\{y_i\}_{i=1}^{N}|\mb{z},\mb{B})]$ can be computed via Eq.(15-17), $\mathbb{E}_{q(\mb{H},\mb{z})}[\log p(\mb{z}|\mb{H})]$ can be computed via Eq.(11-14). $\mathbb{E}_{q(\mb{H})}[\log \pi(\mb{H})]$, $\mathbb{E}_{q(\mb{B})}[\log \pi(\mb{B})]$, $\mathbb{E}_{q(\mb{B})}[\log q(\mb{B})]$, $\mathbb{E}_{q(\mb{H})}[\log q(\mb{H})]$ can be computed in a way similar to Eq.(7). $\mathbb{E}_{q(\mb{z})}[\log q(\mb{z})]$ can be computed via Eq.(18). Given all these expectations, we can get an analytical expression of the objective in Eq.(19) and learn the parameters by optimizing this objective. Regarding how to optimize the mutual angular regularizers $\Omega(\{\mathbb{E}_{q(\tilde{\bs\beta}_k)}[\tilde{\bs\beta}_k]\}_{k=1}^K)$ and $\Omega(\{\mathbb{E}_{q(\tilde{\bs\eta}_k)}[\tilde{\bs\eta}_k]\}_{k=1}^K)$, please refer to \citep{xie2015diversifying} for details.

\vskip 0.2in
\bibliography{egbib}

\begin{thebibliography}{63}
\providecommand{\natexlab}[1]{#1}
\providecommand{\url}[1]{\texttt{#1}}
\expandafter\ifx\csname urlstyle\endcsname\relax
  \providecommand{\doi}[1]{doi: #1}\else
  \providecommand{\doi}{doi: \begingroup \urlstyle{rm}\Url}\fi

\bibitem[Abeywardana(2015)]{Abeywardana}
Sachin Abeywardana.
\newblock Expectation and covariance of von mises-fisher distribution.
\newblock In \emph{https://sachinruk.github.io/blog/von-Mises-Fisher/}, 2015.

\bibitem[Affandi et~al.(2013)Affandi, Fox, and Taskar]{affandi2013approximate}
Raja~Hafiz Affandi, Emily Fox, and Ben Taskar.
\newblock Approximate inference in continuous determinantal processes.
\newblock In \emph{Advances in Neural Information Processing Systems}, pages
  1430--1438, 2013.

\bibitem[Aldous(1985)]{aldous1985exchangeability}
David~J Aldous.
\newblock \emph{Exchangeability and related topics}.
\newblock Springer, 1985.

\bibitem[Banfield et~al.(2005)Banfield, Hall, Bowyer, and
  Kegelmeyer]{banfield2005ensemble}
Robert~E Banfield, Lawrence~O Hall, Kevin~W Bowyer, and W~Philip Kegelmeyer.
\newblock Ensemble diversity measures and their application to thinning.
\newblock \emph{Information Fusion}, 2005.

\bibitem[Bishop(1998)]{bishop1998latent}
Christopher~M Bishop.
\newblock Latent variable models.
\newblock In \emph{Learning in Graphical Models}, pages 371--403. Springer,
  1998.

\bibitem[Bishop and Tipping(2003)]{bishop2003bayesian}
Christopher~M Bishop and Michael~E Tipping.
\newblock Bayesian regression and classification.
\newblock \emph{Nato Science Series sub Series III Computer And Systems
  Sciences}, 190:\penalty0 267--288, 2003.

\bibitem[Blei and Lafferty(2006)]{blei2006correlated}
David Blei and John Lafferty.
\newblock Correlated topic models.
\newblock In \emph{Advances in Neural Information Processing Systems},
  volume~18, page 147. MIT, 2006.

\bibitem[Blei(2014)]{blei2014build}
David~M Blei.
\newblock Build, compute, critique, repeat: Data analysis with latent variable
  models.
\newblock \emph{Annual Review of Statistics and Its Application}, 2014.

\bibitem[Blei et~al.(2003)Blei, Ng, and Jordan]{blei2003latent}
David~M Blei, Andrew~Y Ng, and Michael~I Jordan.
\newblock Latent dirichlet allocation.
\newblock \emph{Journal of Machine Learning Research}, 2003.

\bibitem[Blei et~al.(2006)Blei, Jordan, et~al.]{blei2006variational}
David~M Blei, Michael~I Jordan, et~al.
\newblock Variational inference for dirichlet process mixtures.
\newblock \emph{Bayesian analysis}, 1\penalty0 (1):\penalty0 121--143, 2006.

\bibitem[Bouchard(2007)]{bouchard2007efficient}
Guillaume Bouchard.
\newblock Efficient bounds for the softmax function, applications to inference
  in hybrid models.
\newblock 2007.

\bibitem[Breiman(2001)]{breiman2001random}
Leo Breiman.
\newblock Random forests.
\newblock \emph{Machine Learning}, 45\penalty0 (1):\penalty0 5--32, 2001.

\bibitem[Burges(1998)]{burges1998tutorial}
Christopher~JC Burges.
\newblock A tutorial on support vector machines for pattern recognition.
\newblock \emph{Data Mining and Knowledge Discovery}, 2\penalty0 (2):\penalty0
  121--167, 1998.

\bibitem[Byrne and Girolami(2013)]{byrne2013geodesic}
Simon Byrne and Mark Girolami.
\newblock Geodesic monte carlo on embedded manifolds.
\newblock \emph{Scandinavian Journal of Statistics}, 40\penalty0 (4):\penalty0
  825--845, 2013.

\bibitem[Cai et~al.(2005)Cai, He, and Han]{cai2005document}
Deng Cai, Xiaofei He, and Jiawei Han.
\newblock Document clustering using locality preserving indexing.
\newblock \emph{IEEE Transactions on Knowledge and Data Engineering},
  17\penalty0 (12):\penalty0 1624--1637, 2005.

\bibitem[Cogswell et~al.(2015)Cogswell, Ahmed, Girshick, Zitnick, and
  Batra]{cogswell2015reducing}
Michael Cogswell, Faruk Ahmed, Ross Girshick, Larry Zitnick, and Dhruv Batra.
\newblock Reducing overfitting in deep networks by decorrelating
  representations.
\newblock \emph{arXiv preprint arXiv:1511.06068}, 2015.

\bibitem[Doshi-Velez and Ghahramani(2009)]{doshi2009accelerated}
Finale Doshi-Velez and Zoubin Ghahramani.
\newblock Accelerated sampling for the indian buffet process.
\newblock In \emph{Proceedings of the 26th annual international conference on
  machine learning}, pages 273--280. ACM, 2009.

\bibitem[Fei-Fei et~al.(2007)Fei-Fei, Fergus, and Perona]{fei2007learning}
Li~Fei-Fei, Rob Fergus, and Pietro Perona.
\newblock Learning generative visual models from few training examples: An
  incremental bayesian approach tested on 101 object categories.
\newblock \emph{Computer vision and Image understanding}, 106\penalty0
  (1):\penalty0 59--70, 2007.

\bibitem[Ferguson(1973)]{ferguson1973bayesian}
Thomas~S Ferguson.
\newblock A bayesian analysis of some nonparametric problems.
\newblock \emph{The annals of statistics}, pages 209--230, 1973.

\bibitem[Ghahramani and Griffiths(2005)]{ghahramani2005infinite}
Zoubin Ghahramani and Thomas~L Griffiths.
\newblock Infinite latent feature models and the indian buffet process.
\newblock In \emph{Advances in neural information processing systems}, pages
  475--482, 2005.

\bibitem[Gilks(2005)]{gilks2005markov}
Walter~R Gilks.
\newblock \emph{Markov chain Monte Carlo}.
\newblock Wiley Online Library, 2005.

\bibitem[Gilks and Wild(1992)]{gilks1992adaptive}
Walter~R Gilks and Pascal Wild.
\newblock Adaptive rejection sampling for gibbs sampling.
\newblock \emph{Applied Statistics}, pages 337--348, 1992.

\bibitem[Girolami and Calderhead(2011)]{girolami2011riemann}
Mark Girolami and Ben Calderhead.
\newblock Riemann manifold langevin and hamiltonian monte carlo methods.
\newblock \emph{Journal of the Royal Statistical Society: Series B (Statistical
  Methodology)}, 73\penalty0 (2):\penalty0 123--214, 2011.

\bibitem[Griffiths and Ghahramani()]{griffiths2005infinite}
Thomas Griffiths and Zoubin Ghahramani.
\newblock Infinite latent feature models and the indian buffet process.
\newblock In \emph{Advances in Neural Information Processing Systems}.

\bibitem[Hastings(1970)]{hastings1970monte}
W~Keith Hastings.
\newblock Monte carlo sampling methods using markov chains and their
  applications.
\newblock \emph{Biometrika}, 57\penalty0 (1):\penalty0 97--109, 1970.

\bibitem[Hinton and Salakhutdinov(2006)]{hinton2006reducing}
Geoffrey~E Hinton and Ruslan~R Salakhutdinov.
\newblock Reducing the dimensionality of data with neural networks.
\newblock \emph{Science}, 313\penalty0 (5786):\penalty0 504--507, 2006.

\bibitem[Hjort et~al.(2010)Hjort, Holmes, M{\"u}ller, and
  Walker]{hjort2010bayesian}
Nils~Lid Hjort, Chris Holmes, Peter M{\"u}ller, and Stephen~G Walker.
\newblock \emph{Bayesian nonparametrics}, volume~28.
\newblock Cambridge University Press, 2010.

\bibitem[Hoffman et~al.(2013)Hoffman, Blei, Wang, and
  Paisley]{hoffman2013stochastic}
Matthew~D Hoffman, David~M Blei, Chong Wang, and John Paisley.
\newblock Stochastic variational inference.
\newblock \emph{The Journal of Machine Learning Research}, 14\penalty0
  (1):\penalty0 1303--1347, 2013.

\bibitem[Jaakkola and Jordan(1997)]{jaakkola1997variational}
T~Jaakkola and Michael~I Jordan.
\newblock A variational approach to bayesian logistic regression models and
  their extensions.
\newblock In \emph{Sixth International Workshop on Artificial Intelligence and
  Statistics}, 1997.

\bibitem[Jalali et~al.(2015)Jalali, Xiao, and Fazel]{jalali2015variational}
Amin Jalali, Lin Xiao, and Maryam Fazel.
\newblock Variational gram functions: Convex analysis and optimization.
\newblock \emph{arXiv preprint arXiv:1507.04734}, 2015.

\bibitem[Jordan and Jacobs(1994)]{jordan1994hierarchical}
Michael~I Jordan and Robert~A Jacobs.
\newblock Hierarchical mixtures of experts and the em algorithm.
\newblock \emph{Neural computation}, 6\penalty0 (2):\penalty0 181--214, 1994.

\bibitem[Kindermann et~al.(1980)Kindermann, Snell,
  et~al.]{kindermann1980markov}
Ross Kindermann, James~Laurie Snell, et~al.
\newblock \emph{Markov random fields and their applications}, volume~1.
\newblock American Mathematical Society Providence, 1980.

\bibitem[Knott and Bartholomew(1999)]{knott1999latent}
Martin Knott and David~J Bartholomew.
\newblock \emph{Latent variable models and factor analysis}.
\newblock Number~7. Edward Arnold, 1999.

\bibitem[Koller and Friedman(2009)]{koller2009probabilistic}
Daphne Koller and Nir Friedman.
\newblock \emph{Probabilistic graphical models: principles and techniques}.
\newblock MIT press, 2009.

\bibitem[Kulesza et~al.(2012)Kulesza, Taskar, et~al.]{kulesza2012determinantal}
Alex Kulesza, Ben Taskar, et~al.
\newblock Determinantal point processes for machine learning.
\newblock \emph{Foundations and Trends{\textregistered} in Machine Learning},
  2012.

\bibitem[Kuncheva and Whitaker(2003)]{kuncheva2003measures}
Ludmila~I Kuncheva and Christopher~J Whitaker.
\newblock Measures of diversity in classifier ensembles and their relationship
  with the ensemble accuracy.
\newblock \emph{Machine learning}, 2003.

\bibitem[Lazebnik et~al.(2006)Lazebnik, Schmid, and Ponce]{lazebnik2006beyond}
Svetlana Lazebnik, Cordelia Schmid, and Jean Ponce.
\newblock Beyond bags of features: Spatial pyramid matching for recognizing
  natural scene categories.
\newblock In \emph{Computer Vision and Pattern Recognition, IEEE Computer
  Society Conference on}, volume~2, pages 2169--2178, 2006.

\bibitem[Lewis et~al.(2004)Lewis, Yang, Rose, and Li]{lewis2004rcv1}
David~D Lewis, Yiming Yang, Tony~G Rose, and Fan Li.
\newblock Rcv1: A new benchmark collection for text categorization research.
\newblock \emph{The Journal of Machine Learning Research}, 5:\penalty0
  361--397, 2004.

\bibitem[Lowe(1999)]{lowe1999object}
David~G Lowe.
\newblock Object recognition from local scale-invariant features.
\newblock In \emph{The Proceedings of the Seventh IEEE International Conference
  on Computer Vision}, volume~2, pages 1150--1157. Ieee, 1999.

\bibitem[Lowe(2004)]{lowe2004distinctive}
David~G Lowe.
\newblock Distinctive image features from scale-invariant keypoints.
\newblock \emph{International journal of computer vision}, 60\penalty0
  (2):\penalty0 91--110, 2004.

\bibitem[Malkin and Bilmes(2008)]{malkin2008ratio}
Jonathan Malkin and Jeff Bilmes.
\newblock Ratio semi-definite classifiers.
\newblock In \emph{IEEE International Conference on Acoustics, Speech and
  Signal Processing}, pages 4113--4116. IEEE, 2008.

\bibitem[Mardia and Jupp(2009)]{mardia2009directional}
Kanti~V Mardia and Peter~E Jupp.
\newblock \emph{Directional statistics}, volume 494.
\newblock John Wiley \& Sons, 2009.

\bibitem[Neal(2012)]{neal2012bayesian}
Radford~M Neal.
\newblock \emph{Bayesian learning for neural networks}, volume 118.
\newblock Springer Science \& Business Media, 2012.

\bibitem[Partalas et~al.(2008)Partalas, Tsoumakas, and
  Vlahavas]{partalas2008focused}
Ioannis Partalas, Grigorios Tsoumakas, and Ioannis~P Vlahavas.
\newblock Focused ensemble selection: A diversity-based method for greedy
  ensemble selection.
\newblock In \emph{European Conference on Artificial Intelligence}, 2008.

\bibitem[Platt et~al.(1999)]{platt1999fast}
John Platt et~al.
\newblock Fast training of support vector machines using sequential minimal
  optimization.
\newblock \emph{Advances in Kernel Methods Support Vector Learning}, 3, 1999.

\bibitem[Rasmussen(1999)]{rasmussen1999infinite}
Carl~Edward Rasmussen.
\newblock The infinite gaussian mixture model.
\newblock In \emph{Advances in neural information processing systems},
  volume~12, pages 554--560, 1999.

\bibitem[Teh and Gorur(2009)]{teh2009indian}
Yee~W Teh and Dilan Gorur.
\newblock Indian buffet processes with power-law behavior.
\newblock In \emph{Advances in neural information processing systems}, pages
  1838--1846, 2009.

\bibitem[Teh and Ghahramani(2007)]{tehstick}
Yee~Whye Teh and Zoubin Ghahramani.
\newblock Stick-breaking construction for the indian buffet process.
\newblock 2007.

\bibitem[Wainwright et~al.(2008)Wainwright, Jordan,
  et~al.]{wainwright2008graphical}
Martin~J Wainwright, Michael~I Jordan, et~al.
\newblock Graphical models, exponential families, and variational inference.
\newblock \emph{Foundations and Trends{\textregistered} in Machine Learning},
  1\penalty0 (1--2):\penalty0 1--305, 2008.

\bibitem[Wang and Blei(2013)]{wang2013variational}
Chong Wang and David~M Blei.
\newblock Variational inference in nonconjugate models.
\newblock \emph{Journal of Machine Learning Research}, 14\penalty0
  (1):\penalty0 1005--1031, 2013.

\bibitem[Wang et~al.(2014)Wang, Zhao, Sun, Yan, Wang, Jin, Wang, Gao, Law, and
  Zeng]{wang2014peacock}
Yi~Wang, Xuemin Zhao, Zhenlong Sun, Hao Yan, Lifeng Wang, Zhihui Jin, Liubin
  Wang, Yang Gao, Ching Law, and Jia Zeng.
\newblock Peacock: Learning long-tail topic features for industrial
  applications.
\newblock \emph{ACM Transactions on Intelligent Systems and Technology}, 2014.

\bibitem[Wasserman(2013)]{wasserman2013all}
Larry Wasserman.
\newblock \emph{All of statistics: a concise course in statistical inference}.
\newblock Springer Science \& Business Media, 2013.

\bibitem[Waterhouse et~al.(1996)Waterhouse, MacKay, Robinson,
  et~al.]{waterhouse1996bayesian}
Steve Waterhouse, David MacKay, Tony Robinson, et~al.
\newblock Bayesian methods for mixtures of experts.
\newblock \emph{Advances in Neural Information Processing Systems}, pages
  351--357, 1996.

\bibitem[Wilkinson(2015)]{Wilkinson}
Darren Wilkinson.
\newblock Metropolis hastings mcmc when the proposal and target have differing
  support.
\newblock In
  \emph{https://darrenjw.wordpress.com/2012/06/04/metropolis-hastings-mcmc-when-the-proposal-and-target-have-differing-support/},
  2015.

\bibitem[Xiao et~al.(2010)Xiao, Hays, Ehinger, Oliva, Torralba,
  et~al.]{xiao2010sun}
Jianxiong Xiao, James Hays, Krista Ehinger, Aude Oliva, Antonio Torralba,
  et~al.
\newblock Sun database: Large-scale scene recognition from abbey to zoo.
\newblock In \emph{IEEE Conference on Computer Vision and Pattern Recognition},
  pages 3485--3492. IEEE, 2010.

\bibitem[Xie(2015)]{xie2015learning}
Pengtao Xie.
\newblock Learning compact and effective distance metrics with diversity
  regularization.
\newblock In \emph{European Conference on Machine Learning}, 2015.

\bibitem[Xie et~al.(2015)Xie, Deng, and Xing]{xie2015diversifying}
Pengtao Xie, Yuntian Deng, and Eric~P. Xing.
\newblock Diversifying restricted boltzmann machine for document modeling.
\newblock In \emph{ACM SIGKDD Conference on Knowledge Discovery and Data
  Mining}, 2015.

\bibitem[Xie et~al.(2016)Xie, Zhu, and Xing]{xie2016diversity}
Pengtao Xie, Jun Zhu, and Eric~P. Xing.
\newblock Diversity-promoting bayesian learning of latent variable models.
\newblock In \emph{International Conference on Machine Learning}, 2016.

\bibitem[Yu et~al.(2011)Yu, Li, and Zhou]{yu2011diversity}
Yang Yu, Yu-Feng Li, and Zhi-Hua Zhou.
\newblock Diversity regularized machine.
\newblock In \emph{International Joint Conference on Artificial Intelligence}.
  Citeseer, 2011.

\bibitem[Zhu et~al.(2011)Zhu, Chen, and Xing]{zhu2011infinite}
Jun Zhu, Ning Chen, and Eric~P Xing.
\newblock Infinite svm: a dirichlet process mixture of large-margin kernel
  machines.
\newblock In \emph{Proceedings of the 28th International Conference on Machine
  Learning}, pages 617--624, 2011.

\bibitem[Zhu et~al.(2014{\natexlab{a}})Zhu, Chen, Perkins, and
  Zhang]{zhu2014gibbs}
Jun Zhu, Ning Chen, Hugh Perkins, and Bo~Zhang.
\newblock Gibbs max-margin topic models with data augmentation.
\newblock \emph{Journal of Machine Learning Research}, 15\penalty0
  (1):\penalty0 1073--1110, 2014{\natexlab{a}}.

\bibitem[Zhu et~al.(2014{\natexlab{b}})Zhu, Chen, and Xing]{zhu2014bayesian}
Jun Zhu, Ning Chen, and Eric~P Xing.
\newblock Bayesian inference with posterior regularization and applications to
  infinite latent svms.
\newblock \emph{Journal of Machine Learning Research}, 15\penalty0
  (1):\penalty0 1799--1847, 2014{\natexlab{b}}.

\bibitem[Zou and Adams(2012)]{Zou_priorsfor}
James~Y. Zou and Ryan~P. Adams.
\newblock Priors for diversity in generative latent variable models.
\newblock In \emph{Advances in Neural Information Processing Systems}, 2012.

\end{thebibliography}
\bibliographystyle{natbib}
 
\end{document}